\documentclass[10pt,journal,compsoc]{IEEEtran}
\usepackage{amsmath,amsfonts,bm}
\usepackage{algorithmic}
\usepackage{algorithm}
\usepackage{array}
\usepackage{subcaption}
\usepackage{hyperref}
\usepackage{textcomp}
\usepackage{stfloats}
\usepackage{url}
\usepackage{verbatim}
\usepackage{graphicx}
\usepackage{enumerate}
\usepackage{mathtools}
\usepackage{amsfonts,amssymb,url} 
\usepackage{amsthm}
\usepackage{setspace}
\usepackage{comment}
\usepackage{caption}
\usepackage{multirow}
\usepackage{adjustbox}
\usepackage[english]{babel}

\newcommand{\valpha}{\boldsymbol{\alpha}}
\newcommand{\vbeta}{\boldsymbol{\beta}}

\newtheorem{theorem}{Theorem}
\newtheorem{lemma}{Lemma}
\newtheorem{proposition}{Proposition}

\newtheorem{thm}{Theorem}[section]

\newtheorem{cor}{Corollary}

\newcommand{\Ma}[1]{\ensuremath{\mathbf{#1}}}
\newcommand{\Ve}[1]{\ensuremath{\mathbf{#1}}}

\newcommand{\E}{\mathbb{E}}

\newcommand{\Real}{\ensuremath{\mathbb{R}}}

\newcommand{\R}{\Real}

\DeclareMathOperator{\Trace}{tr}

\newcommand{\norm}[1]{\left\Vert#1\right\Vert}

\newcommand{\mA}{\Ma{A}}

\newcommand{\mI}{\Ma{I}}

\newcommand{\mK}{\Ma{K}}
\newcommand{\mL}{\Ma{L}}
\newcommand{\mM}{\Ma{M}}
\newcommand{\mP}{\Ma{P}}
\newcommand{\mQ}{\Ma{Q}}

\newcommand{\vr}{\Ve{r}}
\newcommand{\vs}{\Ve{s}}
\newcommand{\vu}{\Ve{u}}
\newcommand{\vv}{\Ve{v}}
\newcommand{\vx}{\Ve{x}}
\newcommand{\vy}{\Ve{y}}
\newcommand{\vz}{\Ve{z}}
\newcommand{\vzero}{\Ve{0}}

\newcommand{\calN}{\ensuremath{\mathcal{N}}}

\newcommand{\calR}{\ensuremath{\mathcal{R}}}

\newcommand{\mC}{\Ma{C}}

\newcommand{\mW}{\Ma{W}}
\renewcommand{\Trace}[1]{\mathrm{tr}\left[#1\right]}

\newcommand{\mU}{\Ma{U}}
\newcommand{\mV}{\Ma{V}}

\newcommand{\mX}{\Ma{X}}
\newcommand{\mY}{\Ma{Y}}
\newcommand{\mG}{\Ma{G}}
\newcommand{\mS}{\Ma{S}}
\newcommand{\mH}{\Ma{H}}

\newcommand{\mR}{\Ma{R}}
\newcommand{\citep}{\cite}

\ifCLASSOPTIONcompsoc
  \usepackage[nocompress]{cite}
\else
  \usepackage{cite}
\fi
\begin{document}

\title{Multi-modality fusion using canonical correlation analysis methods: Application in breast cancer survival prediction from histology and genomics}

\author{Vaishnavi Subramanian, Tanveer Syeda-Mahmood, and Minh N. Do
        \thanks{This project has been funded by the Jump ARCHES endowment through the Health Care Engineering Systems Center, University of Illinois at Urbana-Champaign and the IBM-Illinois Center for Cognitive Computing Systems Research (C3SR).
  }
\thanks{V. Subramanian and M. N. Do are with the Electrical and Computer Engineering Department, University of Illinois at Urbana-Champaign, Urbana, IL 61801 USA (e-mail: vs5@illinois.edu; minhdo@illinois.edu)
}
\thanks{T. Syeda-Mahmood is with IBM Almaden Research Center, San Jose, CA 95120 USA (e-mail: stf@us.ibm.com).}

}


\markboth{This work has been submitted to the IEEE for possible publication. Copyright and access may change without notice. }{} 


\IEEEtitleabstractindextext{%
\begin{abstract}
The availability of multi-modality datasets provides a unique opportunity to characterize the same object of interest using multiple viewpoints more comprehensively. 
In this work, we investigate the use of canonical correlation analysis (CCA) and penalized variants of CCA (pCCA) for the fusion of two modalities. We study a simple graphical model for the generation of two-modality data. We analytically show that, with known model parameters, posterior mean estimators that jointly use both modalities outperform arbitrary linear mixing of single modality posterior estimators in latent variable prediction. Penalized extensions of CCA (pCCA) that incorporate domain knowledge can discover correlations with high-dimensional, low-sample data, whereas traditional CCA is inapplicable. To facilitate the generation of multi-dimensional embeddings with pCCA, we propose two matrix deflation schemes that enforce desirable properties exhibited by CCA. We propose a two-stage prediction pipeline using pCCA embeddings generated with deflation for latent variable prediction by combining all the above. On simulated data, our proposed model drastically reduces the mean-squared error in latent variable prediction. When applied to publicly available histopathology data and RNA-sequencing data from The Cancer Genome Atlas (TCGA) breast cancer patients, our model can outperform principal components analysis (PCA) embeddings of the same dimension in survival prediction.
\end{abstract}

 }

 \maketitle

%
\IEEEraisesectionheading{\section{Introduction}\label{sec:introduction}}
\label{sec:intro}

Breast cancer in females is the most commonly diagnosed cancer worldwide. 
It is estimated that the United States alone will witness a striking 284,200 new breast cancer cases and 44,130 deaths from breast cancer in 2021~\citep{siegel2021cancer}. Given this increasing incidence rate, it is crucial to develop improved techniques for patient survival prediction. More accurate survival prediction will aid in clinical decision-making, enabling the right treatment and care to be provided with better outcomes at reduced costs to patients. 

\begin{figure}
    \centering
    \includegraphics[width=0.425\textwidth]{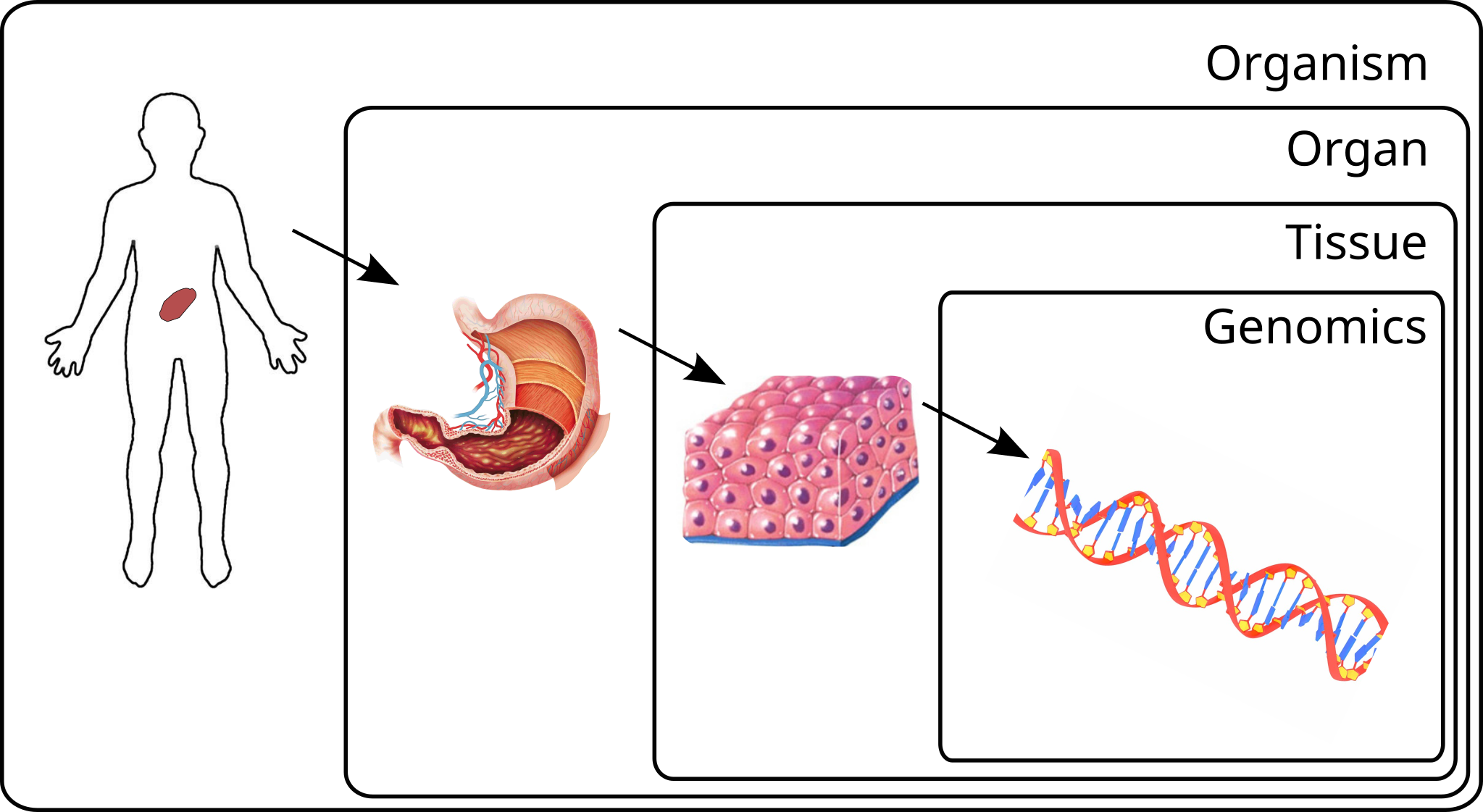}
    \caption{Multi-scale cancer data: Information about the same cancer captured at the organism level (clinical data), the organ level (radiology images), the tissue level (histopathology images) and gene expression levels (RNA-sequencing etc.) should be aggregated together effectively to characterize the underlying cancer. }
    \label{fig:multi-scale-cancer}
\end{figure}

Breast cancer is a highly heterogeneous disease, with heterogeneity manifesting across patients, across tumors of a patient, and also within a single tumor of patients. This makes the accurate survival prediction of breast cancer patients highly challenging
~\citep{rivenbark2013molecular}. A study of the morphological, clinical, and molecular features enables treatment planning in practice. 
These features are captured to varying extents in the different modalities of histology imaging, radiology imaging, genomics and clinical variables~(See Figure~\ref{fig:multi-scale-cancer}). Intelligent aggregation of information from these modalities is essential to yield a complete characterization of the tumor in breast cancer patients. For this, multidisciplinary cancer care teams comprising specialists with focus on different modalities (oncologists, radiologists, pathologists, and nurses) provide expert opinions on the same patient to arrive at the best possible care. Such teams have shown potential in improving patient outcome~\citep{blackwood2020multidisciplinary}.

Algorithms to automatically grade the cancer patients have been developing in the past few decades. One of the first and most widely used methods in breast cancer for survival and prognosis prediction is based on the gene expression of a subset of 50 genes, known as the PAM50 set~\citep{parker2009supervised}. 
The role of broader genomics markers have been studied for survival prediction and therapeutic implications~\citep{zhang2018toward}. 
With the rise of automated image analysis and deep learning methods, several works have addressed the task of survival prediction and response to therapy using histology imaging~\citep{li2019classification}.

With the increasing availability of multi-modality datasets, there is great potential in developing algorithms to grade diseases and further improve patient care by combining information from multiple viewpoints using automated methods. 
In cancer settings, the modalities originate from the same cancer and correspond to histology tissue imaging or radiology imaging, different levels of genomics (such as gene expression and transcriptomics), and overall status of the cancer patient reflected in clinical data (See Figure~\ref{fig:multi-scale-cancer}). 
These modalities jointly describe the cancer in a more comprehensive way, accounting for the cancer properties at different physical scales. 

To take advantage of the diverse information available from the multiple modalities using automated machinery, 
fusion methods have been proposed including methods that combine genomics and clinical data~\citep{gao2021predicting}, different imaging modalities~\citep{mokni2021automatic}, and imaging data with data from other modalities~\citep{golugula2011supervised}. Several works have recently focused on fusing imaging and genomics data using deep learning in the form of convolutional neural networks~\citep{mobadersany2018predicting}, novel fusion modules~\citep{chen2020pathomic,subramanian2020multimodal}, multi-modal autoencoders~\citep{ghosal2021g}, by modelling uncertainty~\citep{wang2021modeling} and machine learning tools~\citep{sun2018integrating,subramanian2021multimodal}. Other methods also integrate information from more than two modalities to capture the most general setting of multi-modality data~\citep{cheerla2019deep,braman2021deep}. 

Cross-correlations across modalities can capture the joint variation of the modalities by identifying which features are positively correlated, which are negatively correlated, and which are uncorrelated. These correlations can be utilized to overcome  missing modalities~\citep{zhou2020brain}, and in learning good representations~\cite{lan2020modality}. 
Statistical methods such as canonical correlation analysis (CCA) and independent components analysis (ICA) can provide an understanding of the cross-interactions and cross-correlations from the same object of interest. Canonical correlation analysis (CCA)~\citep{hotelling1936relations,kettenring1971canonical} identifies correlated linear combinations of two given modalities, or multiple given modalities~\citep{hardoon2004canonical}. Independent components analysis (ICA)~\citep{hyvarinen2000independent} finds statistically independent components, or factors, that compose the multi-modality data assuming non-gaussian data. CCA, ICA, and other methods including group ICA, clustering and multi-factor dimensionality reduction have been widely applied to imaging-genetics and imaging-genomics problems~\citep{liu2014review,liu2021imaging}. Cross-modality correlations and similarities yield joint embeddings with desirable semantic properties~\citep{zolfaghari2021crossclr,radford2021learning} and have also been exploited in recent contrastive learning methods for aligning modalities~\citep{gabeur2020multi}.

In this article, we focus on canonical correlation analysis (CCA) and its penalized variants (pCCA) since these methods make direct use of correlations. 
CCA can highlight the common information shared across two given modalities by identifying the cross-modality correlations. 
Penalties based on prior domain knowledge can be added to the CCA formulation in order to effectively work on high-dimensional, low-sample-size datasets, such as those of cancer imaging-genomics. 
These penalized CCA (pCCA) variants add penalties/constraints in the form of sparsity~\citep{witten2009extensions,parkhomenko2009sparse}, groups~\citep{chen2012efficient} or graphs \citep{du2015gn}. 

The CCA method has been widely used to understand diseases like cancer with multiple modalities, including imaging data~\citep{mokni2021automatic} and spatial transcriptomics data~\citep{stuart2019comprehensive}. In our previous works, we studied the effectiveness of CCA and sparsity-based pCCA in discovering correlations between histology imaging and genomics data~\citep{subramanian2018correlating,subramanian2018integration}.
Despite the utility of CCA in diverse analytical medical settings, the effectiveness of CCA-based embeddings for downstream prediction tasks in the presence of additional labels has not yet been sufficiently investigated. CCA has been used for prediction in computer vision tasks~\citep{sun2005new,haghighat2016fully}, though without rigorous justification. 
Additionally, although CCA embeddings have been used for prediction tasks~\citep{sun2005new,haghighat2016fully}, the use of penalized CCA (pCCA) versions for predictions has remained largely unexplored. 

\begin{figure*}[th]
\begin{subfigure}[b]{0.54\textwidth}
    \centering
    \includegraphics[width=\textwidth]{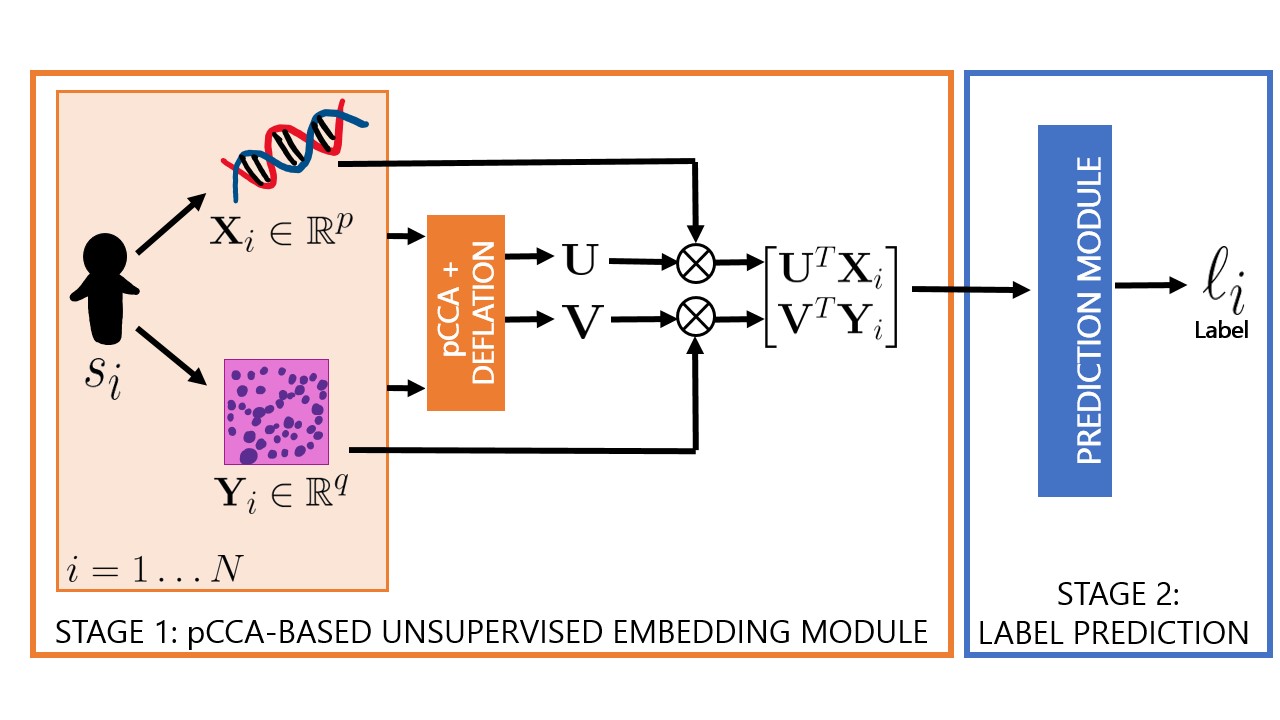}
    \caption{Overview of our proposed two-stage model for prediction. }
    \label{fig:overview}
\end{subfigure}
\hfill
\begin{subfigure}[b]{0.44\textwidth}
    \centering
    \includegraphics[width=\textwidth,trim={0 2cm 0 0},clip]{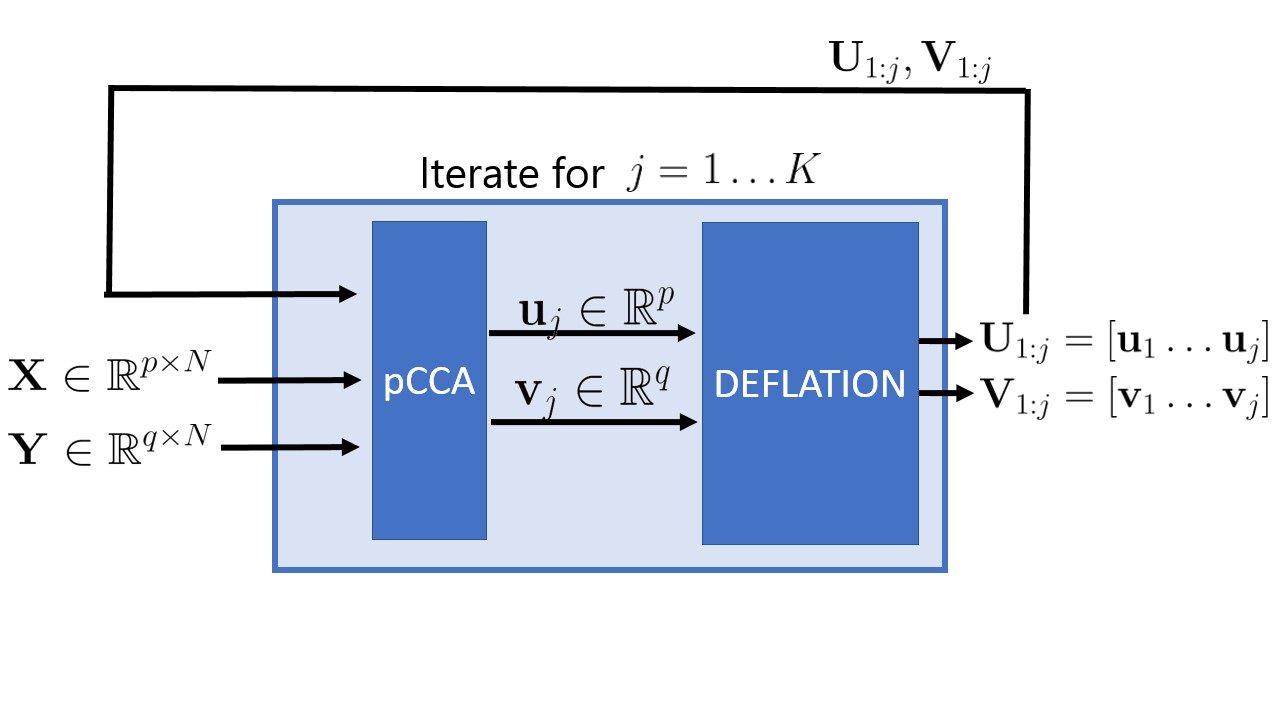}
    \caption{Overview of embedding generation using deflation.} 
    \label{fig:deflation_overview}
\end{subfigure}
\caption{Overview: (a) Our proposed two-stage model for prediction takes in two modalities ($\mX_i$, $\mY_i$) and utilizes pCCA with deflation to generate embeddings ($\tilde{\mX}_i$, $\tilde{\mY}_i$) in an unsupervised manner. The embeddings are concatenated and fed to a (potentially) supervised prediction module for label prediction. (b) To generate embeddings with pCCA, we make use of an iterative deflation scheme where each iteration $j$ identifies canonical weights $\vu_j, \vv_j$. The final embeddings ($\tilde{\mX}_i$, $\tilde{\mY}_i$)  are generated by taking the product $\mU^T \mX_i$ and $\mV^T \mY_i$ where $\mU = \mU_{1:K}= [\vu_1 \dots \vu_K]$ and $\mV = \mV_{1:K} = [\vv_1 \dots \vv_K]$. }
\end{figure*}

\subsection{Contributions}

The main contributions of this work are summarized below. 

\begin{enumerate}[i)]
    \item We analytically show that, under a probabilistic model of two-modality data, the posterior mean estimator of the latent variable that make use of both the modalities together perform better than any arbitrary linear combinations of single modality posterior mean estimators.
    \item We demonstrate how CCA can be used for two-stage prediction based on the above result by recognizing that CCA outputs serves as maximum likelihood estimators of the model parameters in the probabilistic model. Equivalently, we can by-pass the model parameter estimation and directly use the CCA embeddings for prediction as shown in the two-stage pipeline in Figure~\ref{fig:overview}. 
    Stage 1 uses CCA variants to generate multi-dimensional joint embeddings of the two modalities without label supervision. Stage 2 utilizes the joint embeddings to predict latent variables.
    \item We introduce two novel matrix update (deflation) schemes to generate diverse multi-dimensional embeddings with pCCA. These {matrix deflation} schemes, extended from deflation schemes for sparse PCA~\citep{mackey2009deflation}, allow us to capture multi-dimensional correlations by enforcing orthogonality between canonical weights across iterations and, thus, to generate embeddings that capture diverse correlations~(Figure~\ref{fig:deflation_overview}). 
    \item We demonstrate how our fusion module achieves superior performance on simulated data, and can outperform embeddings obtained from principal components analysis (PCA) in TCGA-BRCA survival prediction. 
\end{enumerate}

The rest of our paper is structured as follows. In section~\ref{sec:preliminaries}, we cover the background on CCA and pCCA. Our main contributions are presented in Section~\ref{section:method}, including a mathematical analysis of CCA-based latent variable prediction and our novel deflation schemes. 
Experiments and results are presented in Section~\ref{sec:experiments}. Section~\ref{section:conclusions} concludes our article with key takeaways, limitations of our method, and potential directions of future works.

 \section{Preliminaries}\label{sec:preliminaries}

In this section we set up the mathematical notation, present a simple probabilistic model for the generation of two-modality data, and review CCA and its penalized versions. 

\subsection{Mathematical Notations and Problem Setup}\label{subsec:notations}

\begin{figure}[thb!]
    \centering
    \includegraphics[width=0.13\textwidth]{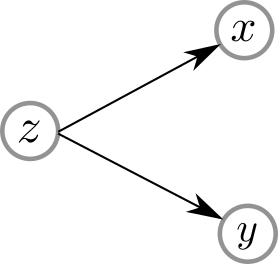}
    \caption{A graphical model for two-modalities~\cite{bach2005probabilistic}. The latent variable of interest ($\vz$) influences the two observed modality variables $\vx$ and $\vy$ (Theorem~\ref{theorem:prob_cca}). For the breast cancer survival prediction setting, $\vz$ is the survival status, $\vx$ the genomics feature and $\vy$ are the imaging features.}
    \label{fig:prob_cca_graphical_model}
\end{figure}

A straightforward probabilistic graphical model for two-modality data is shown graphically in Figure~\ref{fig:prob_cca_graphical_model} and is expressed mathematically as
\[\vz \sim \calN(\vzero, \mI_d), \ \min\{p, q\}>d>1,\]
\[\vx \sim \calN(\mW_x \vz, \Psi_x), \ \mW_x \in \R^{p \times d}, \Psi_x \succcurlyeq 0, \Psi_x \in \R^{p \times p}, \]
\[\vy \sim \calN(\mW_y \vz, \Psi_y), \ \mW_y \in \R^{q \times d}, \Psi_y \succcurlyeq 0, \Psi_y \in \R^{q \times q},\]
where $\vz \in \R^d, \vx \in \R^p, \vy \in \R^q $ and $\mI_d$ denotes the identity matrix of $d$ dimensions. 
The variables $\vx$ and $\vy$ represent two different modalities derived from the underlying latent variable $\vz$ under linear transformations. 
The probabilistic model can be extended to include structural constraints on the combination matrices $\mW_x$ and $\mW_y$.
The above model is a direct extension of the probabilistic PCA model\cite{tipping1999probabilistic} and was used in the probabilistic interpretation of CCA~\cite{bach2005probabilistic}.

Consider $n$ samples from the above model. The $i^{th}$ sample corresponds to an underlying latent variable $\vz_i$ which is observed as a label $\ell_i$, and a single observation of the two modalities as $(\vx_i, \vy_i)$, of dimensions $p$ and $q$ respectively. (For each sample, this corresponds to sampling $\vz$ once, and sampling $\vx, \vy$ once given the sampled $\vz$.)
The observed data from the two modalities 
can be represented as data matrices $\mX \in \R^{p \times n}$ and $\mY \in \R^{q \times n}$, with different samples captured across columns and different features represented across rows. That is, the $i^{th}$ column  of $\mX$, $\mX^i$, and the $i^{th}$ column of $\mY$, $\mY^i$, are both from the same sample, sample $i$, $\forall i \in [1 \dots n]$. The corresponding labels are $\{\ell_i\}_{i=1}^n$. In this work, we will apply our method on breast cancer multi-modality data, with matrix $\mX$ as RNA-sequencing gene expressions, $\mY$ as the cellular and nuclear features from histopathology images, and unseen labels $\ell_i$ corresponding to survival data. Detailed description of these features and labels are in Section~\ref{sec:experiments}.  

\subsection{Canonical Correlation Analysis (CCA)}

\begin{figure*}[htb]
     \centering
     \begin{subfigure}[t]{0.35\textwidth}
         \centering
         \includegraphics[height=0.82\textwidth]{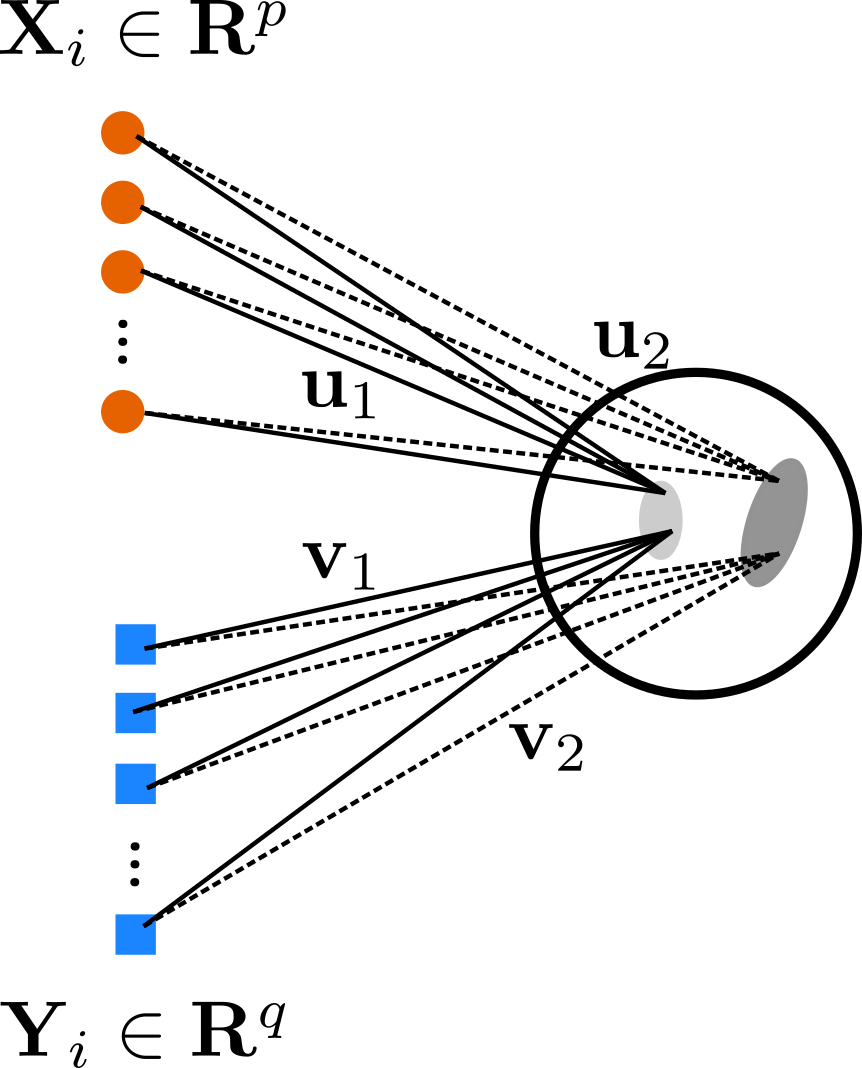}
         \caption{Canonical Correlation Analysis (CCA)}
         \label{fig:cca}
     \end{subfigure}
     \begin{subfigure}[t]{0.3\textwidth}
         \centering
         \includegraphics[height=0.95\textwidth]{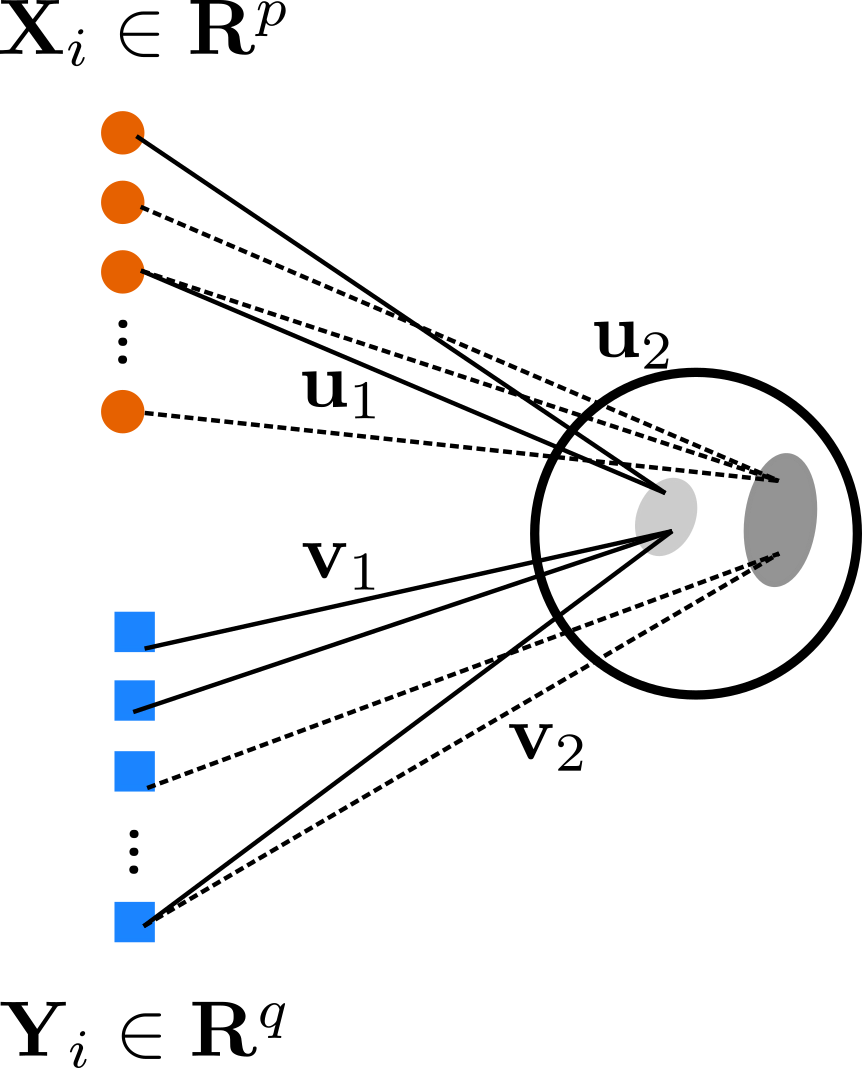}
         \caption{Sparse CCA (SCCA)}
         \label{fig:sparse_cca}
     \end{subfigure}
     \begin{subfigure}[t]{0.3\textwidth}
         \centering
         \includegraphics[height=0.95\textwidth]{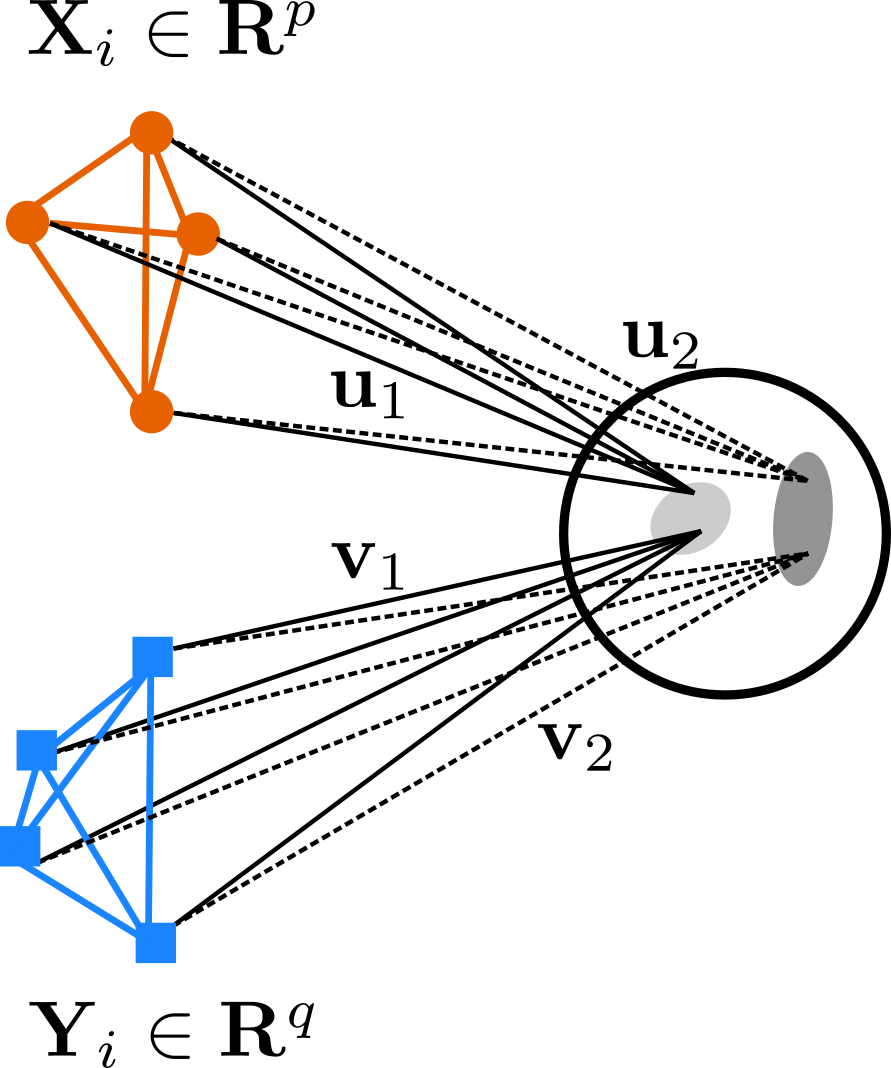}
         \caption{GraphNet SCCA (GN-SCCA)}
         \label{fig:graph_cca}
     \end{subfigure}
        \caption{CCA, SCCA and GN-SCCA aim to identify the first set of canonical weights $\vu_1$ and $\vv_1$ such that the correlation between $\vu^T \mX$ and $\vv^T \mY$ is maximized, subject to no constraints (CCA), sparsity constraints on $\vu, \vv$ (SCCA), and graph-based smoothness constraints on $\vu, \vv$ (GN-SCCA). This can be repeated to identify later sets $(\vu_k, \vv_k), k> 1$ using deflation schemes. Here, the first two sets of canonical weights $(\vu_1, \vv_1$) and $(\vu_2, \vv_2$) are shown. }
        \label{fig:three_ccas}
\end{figure*}

Canonical correlation analysis (CCA)~\cite{hotelling1936relations,kettenring1971canonical}, aims to discover correlations between two sets of variables. The CCA formulation identifies linear combinations of features in each modality such that the resulting vectors are well-correlated, as shown in Figure~\ref{fig:cca}. This is in contrast to traditional correlation analyses which focus on pairwise correlations of individual features across modalities. CCA enables the discovery of underlying concepts that can be represented as linear transformations of the two modalities without requiring any label information. 

Consider zero-mean data ($\Sigma_{i=1}^n \mX^i = 0$, $\Sigma_{i=1}^n \mY^i = 0$, where $\mX^i, \mY^i$ are the $i^{th}$ columns of $\mX$ and $\mY$ respectively). The empirical cross-correlation matrix is $\mC_{xy} = \mX \mY^T$ and the auto-correlation matrices are $\mC_{xx} = \mX \mX^T, \mC_{yy} = \mY \mY^T$. The first iteration of the CCA problem finds 
\begin{equation}\label{eq:CCA}
\rho^* = \max_{\vu, \vv} \vu^T \mC_{xy} \vv \quad \text{s.t.  } \quad \vu^T \mC_{xx} \vu = 1, \vv^T \mC_{yy} \vv = 1,
\end{equation}
with optimal canonical weights ($\vu^*, \vv^*$). The corresponding canonical variates are $(\vu^*)^T\mX$ and $(\vv^*)^T\mY$. This problem is usually simplified to a generalized eigenvalue decomposition \citep{hardoon2004canonical}
using the Lagrangian stationarity of the optimal canonical weights. Another equivalent formulation is 
\begin{equation*}\label{eq:SVD-CCA}
\rho^* = \max_{\tilde{\vu}, \tilde{\vv}} \tilde{{\vu}}^T \mA \tilde{\vv} \ \text{s.t.} \ \tilde{\vu}^T \tilde{\vu} = 1, \ \tilde{\vv}^T \tilde{\vv} = 1,
\end{equation*}
with $\mA = \mC_{xx}^{-1/2} \mC_{xy} \mC_{yy}^{-1/2}$, assuming invertibility of the auto-correlation matrices $\mC_{xx}, \mC_{yy}$.
The solution to this optimization problem is the same as finding the first singular vectors ($\tilde{\vu}^*, \tilde{\vv}^*$) using the singular value decomposition (SVD) of $\mA$~\citep{parkhomenko2009sparse}.
The optimal CCA canonical weights $\vu^*$ and $\vv^*$ can then be obtained as $\vu^* = \mC_{xx}^{-1/2} \tilde{\vu}^*$, $\vv^* = \mC_{yy}^{-1/2} \tilde{\vv}^*$.

The CCA formulation can be further extended~\citep{hardoon2004canonical} to identify $k$ canonical weights $\{\vu_i\}_{i=1}^k$, $\{\vv_i\}_{i=1}^k$ such that the resulting matrices $\mU = [\vu_1 \dots \vu_k]$ and $\mV = [\vv_1 \dots \vv_k]$ are the optimizers for the trace of $\mU^T \mC_{xy} \mV$, $\Trace{\mU^T \mC_{xy} \mV}$, as below
\begin{align*}
     \max_{\mU, \mV} \Trace{\mU^T \mC_{xy} \mV} \ 
    \text{s.t. } &\mU^T \mC_{xx} \mU = \mI_p, \mV^T \mC_{yy} \mV = \mI_q, \nonumber \\
    &    \vu_i^T\mC_{xy} \vv_j = 0, \ \forall \ j\neq i, 
\end{align*} 
and are constrained to be orthogonal across iterations with respect to the empirical correlation matrices.
This optimization problem identifies multiple directions of correlations between the two modalities' features, capturing more connections between the two modalities. In this formulation, the canonical weights $\{\vu_i\}_{i=1}^k$, $\{\vv_i\}_{i=1}^k$ 
are computed simultaneously.

An iterative formulation of SVD performs 
\begin{align*}
\rho_j^* = \max_{\vu_j, \vv_j}& \  \vu_j^T \mC_{xy} \vv_j \\
\text{s.t. } & \vu_j^T \mC_{xx} \vu_j = \vv_j^T \mC_{yy} \vv_j = 1,  \nonumber  \\
& \vu_j^T \mC_{xx} \vu_i = \vv_j^T \mC_{yy} \vv_i = \vu_j^T \mC_{xy} \vv_i = 0, \ \forall i < j. \nonumber
\end{align*}
at the $j^{th}$ iteration, for $j=1 \dots k$.

Since the CCA solution can be derived from the SVD solution for $\mA = \mC_{xx}^{-1/2} \mC_{xy} \mC_{yy}^{-1/2}$, we take a look at the equivalent formulations of SVD for any matrix $\mA$.
%
The simultaneous formulation of SVD of $\mA$ is given by
\begin{align*}
    (\mU^*, \mV^*) &= \arg \max_{\tilde{\mU}, \tilde{\mV}} \Trace{\tilde{\mU}^T \mA \tilde{\mV}}\  \\ 
    &\text{s.t. } \tilde{\mU}^T \tilde{\mU} = \mI_p, \tilde{\mV}^T \tilde{\mV}  = \mI_q. \nonumber
\end{align*}
The iterative formulation of SVD of $\mA$ with Hotelling deflation (See Figure~\ref{fig:hd}) is
\begin{align*}
(\tilde{\vu}_j, \tilde{\vv}_j) &= \arg \max_{\vu, \vv} \vu^T \mA_j \vv \quad \text{s.t.} \ \vu^T\vu = \vv^T \vv = 1, \\
\mA_{j+1} &= \mA_j - \sigma_j \tilde{\vu}_j \tilde{\vv}_j^T, \ \quad \text{(Hotelling Deflation)} \\
\tilde{\mU} &= [\tilde{\vu}_1 \dots \tilde{\vu}_k], \tilde{\mV} = [\tilde{\vv}_1 \dots \tilde{\vv}_k], 
\end{align*}, where $j = 1, 2, \dots k-1$ and $\mA_1 = \mA$.
Both the above formulations are equivalent, up to permutation of columns, as can be proven by the orthonormality conditions of the weight vectors. 
Therefore, carrying forward the result to CCA, the iterative and simultaneous formulations of CCA result in equivalent solutions. 

The iterative formulation of CCA with Hotelling deflation scheme~\citep{hotelling1933analysis} can be used to obtain the the canonical weights by performing the following updates at the end of each SVD iteration: 
\begin{align}\label{eq:hotelling_update}
    \mA_{j+1} = \mA_j - \rho_j \tilde{\vu}_j \tilde{\vv}_j^T, \ \ j = 1, 2, \dots k-1, 
\end{align}
with $\mA_1 = \mC_{xx}^{-1/2}\mC_{xy}\mC_{yy}^{-1/2}$, with singular vectors $\tilde{\vu}_j, \tilde{\vv}_j$ and singular value $\rho_j$. From these, the canonical weights $\vu_j$ and $\vv_j$ can be computed as $\vu_j = \mC_{xx}^{-1/2} \tilde{\vu}_j$, $\vv_j = \mC_{yy}^{-1/2} \tilde{\vv}_j$ $\forall \ j \in [1 \dots k]$.
Since both the iterative and simultaneous formulations of CCA are equivalent, the iterative formulation is preferred because it is computationally cheaper, solving a simpler problem at each iteration. Therefore, we focus on the iterative formulation of CCA. 

\subsection{Latent variable prediction using CCA}\label{subsec:latent_prediction_CCA}

CCA, proposed as an analysis tool, also has probabilistic implications as noted by Bach and Jordan~\cite{bach2005probabilistic}. The authors proved that for the two-modality data model in Section~\ref{subsec:notations}, CCA lends the maximum likelihood estimator of the model parameters. 
  
\begin{theorem}[CCA as ML estimators, Theorem 2~\cite{bach2005probabilistic}]\label{theorem:prob_cca}
Consider the probabilistic model for two-modality data in Section~\ref{subsec:notations}. Let $\mC_{xx}$, $\mC_{yy}$ and $\mC_{xy}$ denote the sample correlation matrices. 
Then, any maximum likelihood (ML) estimator of the model parameters $\mW_x, \mW_y, \Psi_x, \Psi_y$ are of the form
\begin{align*}
   & \widehat \mW_x = \mC_{xx} \mU~\mM_x, \hfill
    & \widehat \mW_y = \mC_{yy} \mV~\mM_y, \\
   & \widehat \Psi_x = \mC_{xx} -\widehat \mW_x \widehat \mW_x^T, \hfill
    & \widehat \Psi_y = \mC_{yy} - \widehat \mW_y\widehat \mW_y^T, 
\end{align*}
where  $\mU \in \R^{p \times d}$ and $\mV \in \R^{q \times d}$  are the canonical weights from CCA stacked column-wise, matrices $\mM_x, \mM_y \in \R^{d \times d}$ are such that $\mM_x \mM_y^T = \mP$, where $\mP \in \R^{d \times d}$ is the diagonal matrix of the first $d$ canonical correlations. 

\end{theorem}
The above theorem proves that the model parameters $\mW_x$, $\mW_y$ with the maximum likelihood can be estimated using the canonical weight matrices $\mU$ and $\mV$ from CCA. These estimated model parameters can then be used to compute posterior estimates of the latent variable $\vz$ via $\E[\vz|\vx], \E[\vz|\vy]$ and $\E[\vz|(\vx, \vy)]$,
with 
\begin{align}
    \mathbb{E}[\vz|\vx] &= \mM_x^T \mU^T \vx = \widehat{\mW}_x^T \mC_{xx}^{-1} \vx, \label{z_given_x} \\
    \mathbb{E}[\vz|\vy] &= \mM_y^T \mV^T \vy = \widehat{\mW}_y^T \mC_{yy}^{-1} \vy, \label{z_given_y}  \\
    \mathbb{E}[\vz | (\vx, \vy)] &= \begin{bmatrix}
        \mM_x \\ \mM_y
    \end{bmatrix}^T \begin{bmatrix}
    \mI & \mP \\
    \mP & \mI
    \end{bmatrix}^{-1}
     \begin{bmatrix}
    \mU^T \vx \\
    \mV^T \vy
    \end{bmatrix} \label{z_given_xy_embedding} \\
    &= \widehat{\mW} \begin{bmatrix}
    \mC_{xx} & \mC_{xy} \\
    \mC_{yx} & \mC_{yy}
\end{bmatrix}^{-1} \begin{bmatrix}
        \vx \\ \vy 
    \end{bmatrix}. \label{z_given_xy_estimator} 
\end{align}

\subsection{Penalized variants of CCA}\label{subsec:pCCA}

The original CCA formulation requires large sample sizes for high-dimensional data ($n~\geq~\max(p,q)$) to successfully evaluate the solution via decompositions like SVD. 
This is restrictive for imaging-genomic studies since most imaging and genomic data are high dimensional with $\max(p,q)~\gg~n$. In such settings, solving the corresponding generalized eigenvalue problem~\cite{hardoon2004canonical} is not computationally feasible, and the auto-correlation matrices $\mC_{xx}$ and $\mC_{yy}$ fail to be invertible such that the SVD-based algorithm fails.

To resolve this, convex penalty constraints $P_x(\vu)$ and $P_y(\vu)$ on canonical weights $\vu$ and $\vv$ respectively can be added~\citep{witten2009extensions,parkhomenko2009sparse,chen2012efficient,du2015gn}. These penalized variants of CCA, henceforth \textit{penalized CCA} or \textit{pCCA}, result in biconvex/bilinear problems which can be solved to a local optimum. 
Adding the convex penalty constraints reduces the search space of solutions and enables CCA to work with high-dimensional, low-sample-size data.
For example, sparsity constraints on canonical weight pair ($\vu, \vv$) result in the Sparse CCA (SCCA) formulation~\cite{witten2009penalized} illustrated in Figure~\ref{fig:sparse_cca}. 
Setting both penalties to be the $\ell_1$ norm, SCCA solves
\begin{align*}
\rho^* = \max_{\vu,\vv} &\ \vu^T \mC_{xy} \vv \\
\nonumber \text{s.t.}& \norm{\vu}_2\leq1,\norm{\vv}_2 \leq1, \norm{\vu}_1\leq c_1, \norm{\vv}_1\leq c_2,
\end{align*}
where $\mC_{xy} = \mX \mY^T.$ 
This biconvex problem is solved in an alternating manner using soft-thresholding based update rules~\citep{witten2009extensions,witten2009penalized}.
Note that the unit-norm constraints here are on the canonical weights $\vu$ and $\vv$, instead of the canonical variates $\vu^T \mX$ and $\vv^T \mY$ in~\eqref{eq:CCA}, to ensure strict convexity of constraints~\citep{gross2015collaborative}.
Relaxing the unit norm constraints from $\vu^T \mX, \vv^T \mY$ to $\vu, \vv$ in the original CCA problem in~\eqref{eq:CCA} (or equivalently assuming $\mC_{xx} = \mI_p, \mC_{yy} = \mI_q$) results in the SVD problem of the cross-covariance matrix $\mC_{xy}$. 

Other penalties capturing structure in the canonical weights $\vu$ and $\vv$ can be imposed in the pCCA formulation for incorporating dependencies and prior domain knowledge. For example, the \textit{fused-lasso} penalty
to encourage smoothness in the linear combination~\cite{witten2009extensions} and  the group-lasso penalty to 
to encourage non-overlapping groups of features to be in or out of the correlation together~\cite{chen2012efficient} .

Graph-based penalties as a generalization of group-based penalties have also been proposed~\cite{du2015gn}. Here, each modality has an underlying graph structure 
and the graph-constrained elastic net penalties encourage pCCA to select graph-local regions or communities to arrive at Graph-Net SCCA, GN-SCCA (Figure~\ref{fig:graph_cca}).
%
For nodes connected by highly weighted edges, the canonical coefficients should be similar. 
The graphs considered in genomic settings could be from pathways, gene regulatory networks, or protein-protein interaction networks. 
%
%
The resulting graph-based optimization problem is bi-convex and can be solved using alternating optimization in $\vu$ and $\vv$ to a local optimum, similar to SCCA. 

The pCCA variants are well-suited for our application of high-dimensional, low-sample cancer data and additionally provide the opportunity to integrate prior knowledge and structural expectations. In order to generate multiple canonical weights $\vu_1 \dots \vu_k, \vv_1 \dots \vv_k$, the Hotelling deflation scheme is frequently used to deflate the cross-correlation matrix, and repeat the same set of steps. The direct use of Hotelling deflation might fail to learn novel weights across iterations since orthogonality constraints on canonical weights across iterations are no longer enforced with pCCA. Thus, there is a need to develop new methods of generating multiple canonical weights with pCCA.

To recapitulate, in this section we presented a simple probabilistic graphical model for generation of two-modality data, the background of CCA and penalized CCA variants, and how CCA can determine the model parameters for the graphical model of interest.
 \section{Methods}\label{section:method}

In this section, we first mathematically show the value of working with both the available modalities jointly in latent variable prediction. 
We then present two equivalent two-stage approaches to utilize CCA for latent variable prediction - one based on estimation of the model parameters using CCA prior to prediction, and the other based on the direct use of the canonical weights from CCA to generate two-modality embeddings for input to the predictors.
Lastly, we extend the second two-stage approach to work with pCCA embeddings for latent variable prediction in the most general case. To enforce desirable properties in the generation of pCCA embeddings, we introduce two novel matrix deflation schemes. 


\subsection{Latent variable prediction}

Consider the two-modality data model, with the covariance matrices as $\Psi_x = \sigma_x^2 \mI_p$ and $\Psi_y = \sigma_y^2 \mI_q$. Assume that the model parameters $\mW_x, \mW_y, \sigma_x$ and $\sigma_y$ are all known. Let $\hat\vz_c= \E[\vz|(\vx, \vy)]$ be posterior mean estimator which estimates $\vz$ using the two modalities jointly and  $\hat\vz_\beta = \beta \E[\vz|\vx] + \Bar\beta \E[\vz|\vy]$ for $\beta \in [0,1], \Bar\beta = 1-\beta$ be any $\beta$ linear combination of the estimates based on single modalities $\vx$ and $\vy$.
\begin{theorem}\label{theorem_multimodal}
Under the assumed data model with known model parameters $\mW_x, \mW_y, \sigma_x$ and $\sigma_y$, $\hat\vz_c$ is the better estimator compared to $\hat\vz_\beta$ in terms of $e(\hat\vz) = \E_{(\vx, \vy)|\vz}[\norm{\hat\vz - \vz}_2^2]$, the mean squared error in the estimation of $\vz$, with
\[e(\hat\vz_\beta) \geq e(\hat\vz_c) \quad \forall \beta \in [0,1]. \]
\end{theorem}

The above theorem states that the posterior mean estimator of $\vz$ which uses both modalities is better than any arbitrary linear combination of the posterior mean estimators constructed using single modalities. Below, we highlight the key steps and results in the proof. For the complete proof, see Appendix A. 

If only a single modality $\vx$ or $\vy$ is used for the estimation, we have the posterior mean estimators as
\begin{align*}
    \hat{\vz}_x &\triangleq \E [\vz|\vx] 
    = \underbrace{(\mW_x^T \Psi_x^{-1} \mW_x + \mI_d)^{-1} \mW_x^T \Psi_x^{-1}}_{\mG_x} \vx = \mG_x \vx, \\
    \hat{\vz}_y &\triangleq \E [\vz|\vy] = \underbrace{(\mW_y^T \Psi_y^{-1} \mW_y + \mI_d)^{-1} \mW_y^T \Psi_y^{-1}}_{\mG_y} \vy = \mG_y \vy, 
\end{align*}
where $\Psi_x = \sigma_x^2 \mI_p$ and $\Psi_y = \sigma_y^2 \mI_q$. 
A $\beta$-weighted linear mixing of these estimators is 
\begin{align*}
    \hat{\vz}_{\beta} &\triangleq \beta \hat{\vz}_x +  \Bar{\beta} \hat{\vz}_y = \beta \mG_x \vx + \Bar\beta \mG_y \vy \nonumber \\
    &= \underbrace{\begin{bmatrix}
    \mG_x & \mG_y 
    \end{bmatrix}\begin{bmatrix}
    \beta \mI_p & \vzero \\ \vzero & \Bar\beta \mI_q
    \end{bmatrix}}_{\mG_\beta} \begin{bmatrix}
    \vx \\ \vy
    \end{bmatrix} 
    = \mG_\beta \begin{bmatrix}
    \vx \\ \vy
    \end{bmatrix},
\end{align*}
where $\beta \in [0,1]$ and $\Bar{\beta} = 1 - \beta$.
Lastly, the modalities can be used jointly in the estimation of $\vz$ which leads to the estimator that combines $\vx$ and $\vy$ as
\begin{align*}
    \hat{\vz}_c &\triangleq \E[\vz | (\vx, \vy)] \\
    &= \underbrace{(\mW^T \Psi^{-1} \mW + \mI_d)^{-1} \mW^T \Psi^{-1}}_{\mG_c} \begin{bmatrix}
    \vx \\
    \vy
    \end{bmatrix} = \mG_c \begin{bmatrix}
    \vx \\ \vy
    \end{bmatrix},
\end{align*}
where $\mW = \begin{bmatrix}
\mW_x \\ \mW_y
\end{bmatrix}$ and $\Psi = \begin{bmatrix}
\Psi_x & \vzero \\ \vzero & \Psi_y
\end{bmatrix} =  \begin{bmatrix}
\sigma_x^2 \mI_p & \vzero \\ \vzero & \sigma_y^2 \mI_q
\end{bmatrix}$.
\begin{lemma}\label{lemma:error_expression_main}
For any estimator of the form $\hat\vz= \mG \begin{bmatrix}
\vx \\ \vy
\end{bmatrix}$ with $\mG \in \R^{d\times (p+q)}$, 
the error in the estimation of $\vz$ is
\begin{align*}
    e(\hat\vz) &= \vz^T (\mG \mW - \mI_d)^T (\mG \mW - \mI_d) \vz + \Trace{\mG \Psi \mG^T}.
\end{align*}
\end{lemma}
\begin{lemma}\label{lemma:aboutK_main}
Let $\mK$ be given by $\mK = \mG \mW - \mI_d$ where $\mG~=~(\mW^T \Psi^{-1} \mW + \mI_d)^{-1} \mW^T \Psi^{-1} $ and $\Psi \succ 0$. Then, $\mK = -(\mW^T \Psi^{-1} \mW + \mI_d)^{-1}$ and is negative definite. 
\end{lemma}


From Lemma~\ref{lemma:error_expression_main} and ~\ref{lemma:aboutK_main}, we obtain
\begin{align*}
    e(\hat{\vz}_\beta) &=  \vz^T\mK_\beta^T \mK_\beta \vz  + \Trace{\mG_\beta \Psi \mG_\beta^T}, \ \text{and}\\
    e(\hat{\vz}_c) 
    &=  \vz^T\mK_c^T \mK_c \vz  + \Trace{\mG_c \Psi \mG_c^T}, 
\end{align*}
where $\mK_\beta = \mG_\beta \mW - \mI_d$ and $\mK_c = \mG_c \mW - \mI_d$. 
It can be shown that $\mK_\beta - \mK_c \preceq 0$ and $\mK_\beta + \mK_c \preceq 0$. 
These together also imply that $\Trace{\mG_\beta^T \mG_\beta - \mG_c^T \mG_c} \geq 0$. Together, we have
\begin{align*}
    &e(\hat\vz_\beta) - e(\hat\vz_c) \\
    &= \vz^T (\mK_\beta^T \mK_\beta - \mK_c ^T \mK_c) \vz \nonumber  + \Trace{\Psi ( \mG_\beta^T \mG_\beta - \mG_c^T \mG_c )} \\
    &\geq 0.
\end{align*}

\subsection{Latent variable prediction with CCA and pCCA}\label{subsec:prob_cca}

When the model parameters are known, one can compute the posterior mean estimates using the two-modality observations. CCA provides the maximum likelihood estimators for the model parameters in the general case when the model parameters are known, as shown in Section~\ref{subsec:latent_prediction_CCA}. The outputs of CCA can then be used for the posterior mean estimation of $\vz$ as shown in~\eqref{z_given_x}-\eqref{z_given_xy_embedding}.

The equations~\eqref{z_given_x}-\eqref{z_given_xy_embedding} for the latent variable prediction can be cast as two different two-stage prediction models:
\begin{enumerate}
    \item Stage 1: Estimation of the probabilistic model parameters $\widehat{\mW}_x, \widehat{\mW}_y, \widehat{\Psi}_x,\widehat{\Psi}_y$; Stage 2: Posterior mean estimation of $\vz$ based on analytical formula using estimates of the model parameters $\widehat{\mW}$, and 
    \item Stage 1: Estimation of the CCA-based embeddings $\mU^T \vx, \mV^T \vy$; Stage 2: Estimation of $\vz$ using the concatenated vector $\begin{bmatrix} \mU^T \vx \\ \mV^T \vy \end{bmatrix}$. 
    The equations~\eqref{z_given_x}-\eqref{z_given_xy_embedding} lead to a prediction with the matrix $\begin{bmatrix} \mM_x \\ \mM_y \end{bmatrix}$.
\end{enumerate}
The second two-stage prediction model can be adapted to more general settings, when the exact prediction matrices are hard to compute, for example with pCCA. Since we are motivated by cancer imaging-genomics settings which have high-dimensionality low-sample data, the use of pCCA is necessary. This two-stage model can also incorporate a broader range of complex prediction modules, for example supervised predictors like multi-layer perceptrons and random forests, with the CCA embeddings as inputs. Hence, we work with this two-stage model for latent variable prediction, as demonstrated in Figure~\ref{fig:overview}. Adapting this two-stage pipeline to pCCA requires the generation of informative multi-dimensional pCCA embeddings with a wide range of penalties, which we investigate next.


\begin{figure}[h]

    \begin{subfigure}[c]{0.48\textwidth}
    \centering
    \includegraphics[height=0.36\textwidth]{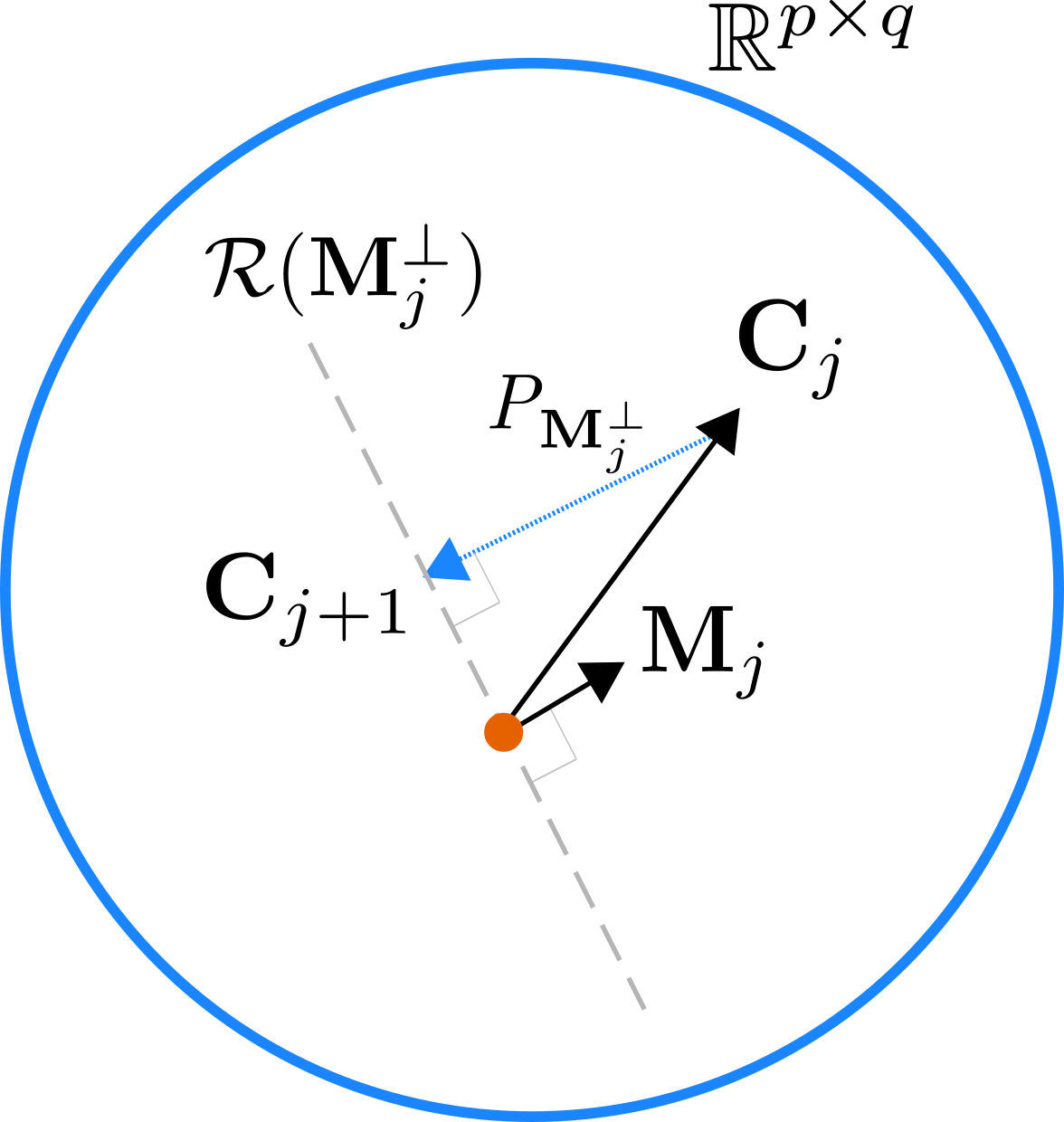}
    \caption{Hotelling Deflation: At iteration $j$ a component $\rho_j$ of the matrix $\mM_j = \vu_j \vv_j^T$ is subtracted from the current cross-covariance matrix $\mC_j$ to obtain $\mC_{j+1}$. In the case of CCA, this results in an orthogonal projection onto the space $\calR(\mM_j^{\perp})$.}
    \label{fig:hd}
    \end{subfigure}
    \par\bigskip
     \begin{subfigure}[c]{0.48\textwidth}
    \centering
    \includegraphics[height=0.36\textwidth]{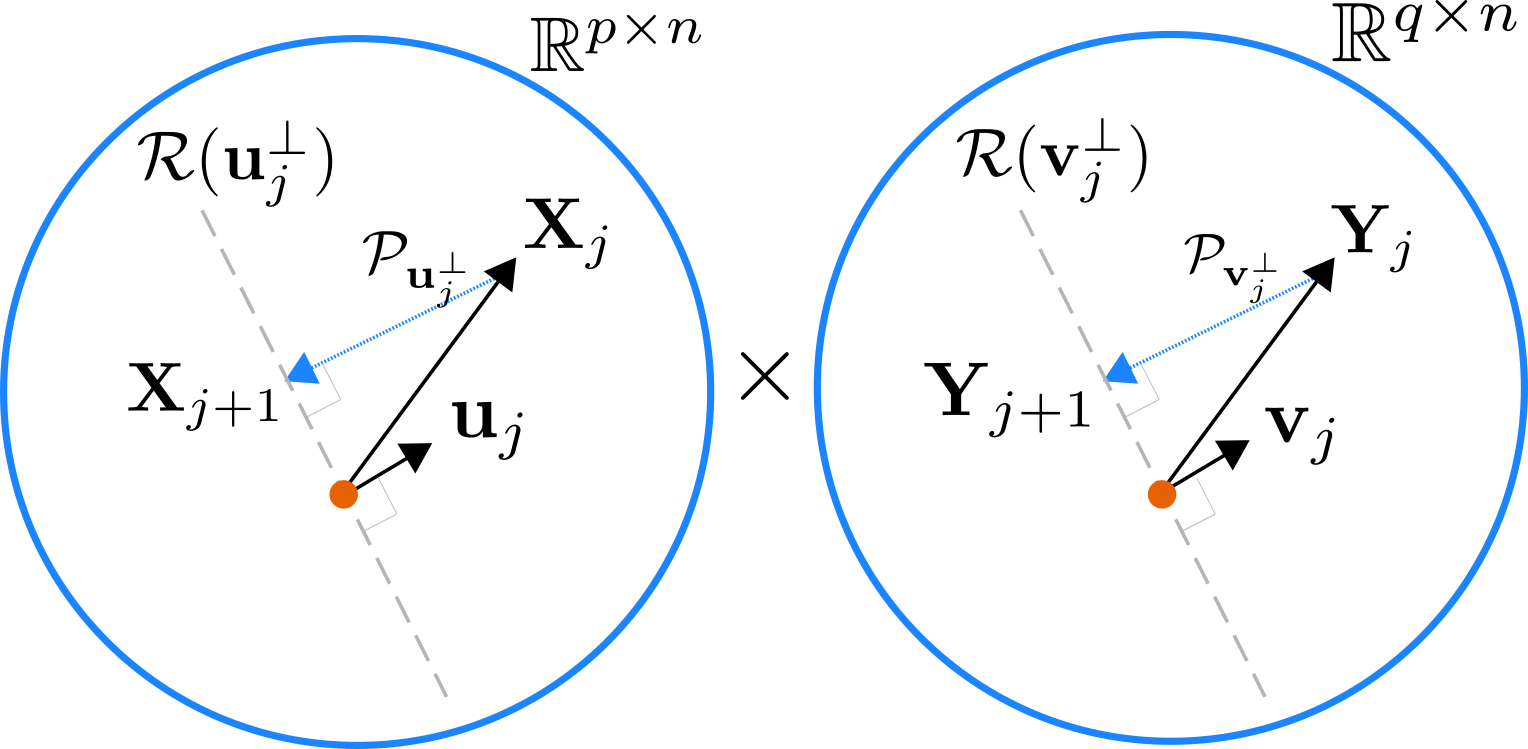}
    \caption{Projected Deflation: At iteration $j$, the data matrices $\mX_j$ and $\mY_j$ are projected onto the orthogonal space of the just-found canonical weights $\vu_j$ and $\vv_j$ respectively. }
    \label{fig:pd}
    \end{subfigure}
    \par\bigskip
     \begin{subfigure}[c]{0.48\textwidth}
    \centering
    \includegraphics[height=0.36\textwidth]{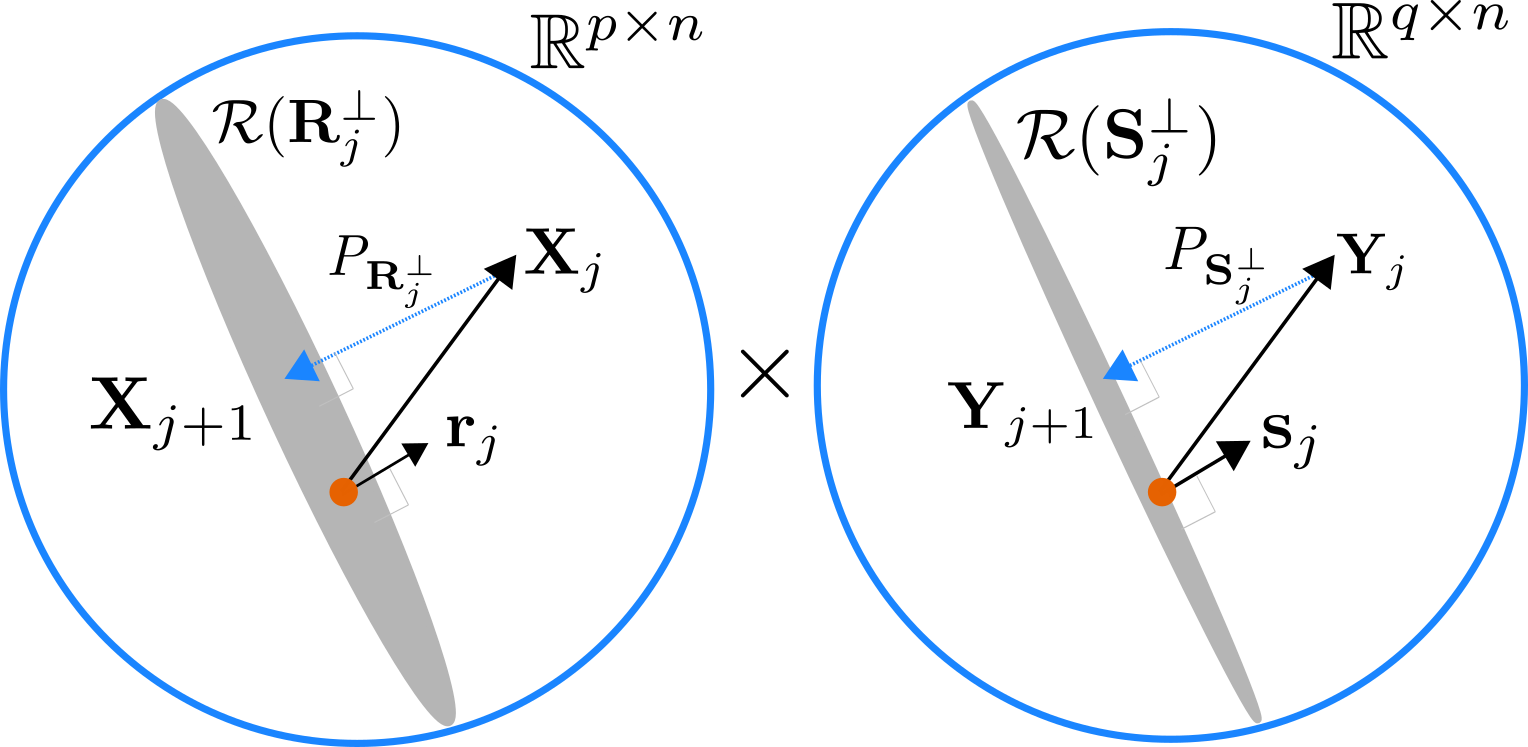}
    \caption{Orthogonalized Projected Deflation: At iteration $j$, the data matrices $\mX_j$ and $\mY_j$ are projected onto the orthogonal space of all the previous found canonical weights. Since vectors $\vr_j$ and $\vs_j$ are the \textit{new} components of the just-found canonical weights $\vu_j$ and $\vv_j$ respectively, it suffices to project the data matrices orthogonal to the vectors $\vr_j$ and $\vs_j$ respectively.}
    \label{fig:opd}
    \end{subfigure}
    \caption{Illustrations of the three deflation schemes used}
\end{figure}

\subsection{Unsupervised embedding generation with pCCA} 
To generate $k$-dimensional embeddings, SCCA~\citep{witten2009penalized,witten2009extensions} adopts the same Hotelling deflation (HD) scheme~\citep{hotelling1933analysis}~(Figure~\ref{fig:hd}) as that used in CCA and SVD~\eqref{eq:hotelling_update}. Specifically, the $j^{th}$ canonical weights of SCCA are identified using
\begin{align}
(\vu_j, \vv_j) =& \arg \max_{\vu,\vv } \ \vu^T \mC_{j-1} \vv \nonumber \\
\nonumber &\text{s.t.} \norm{\vu}_2\leq1,\norm{\vv}_2 \leq1, \norm{\vu}_1 \leq c_1, \norm{\vv}_1\leq c_2.\\
\mC_j =& \mC_{j-1} - \vu_j \vu_j^T \mC_{j-1} \vv_j \vv_j^T \,  \label{HotellingDeflationSCCA}
\end{align}
for $j \geq 1$, where $\mC_0 = \mX \mY^T$. 
The applicability of HD to CCA relies on the close relation between CCA and SVD, and the properties of SVD as a decomposition with orthonormal singular vectors. It is desired to discover novel correlations at each iteration. To support this, CCA + HD  guarantees that CCA will not pick canonical variates already identified in previous iterations, by enforcing that
\begin{align*}
    &\mC_j^T \vu_i = 0, \ \forall i \leq j, 
    &\mC_j \vv_i = 0, \ \forall i \leq j.
\end{align*}
Since the objective $\vu^T \mC_j \vv=0$ for all pairs of vectors $\{(\vu_i, \vv)\}_{i=1}^k$, and $\{(\vu, \vv_i)\}_{i=1}^k$, this condition enforces matrix-based orthogonality on the canonical weights across iterations and disallows the canonical weights to be repeated. This encourages a different correlation (and thus different information) to be identified in later iterations.

However, the above result relies entirely on the relation of the CCA canonical weights with the singular vectors and does not directly carry forward to a generic pCCA+HD. In particular, whenever the canonical weights are different from the singular vectors, the above result does not hold.
The presence of non-zero $\mC_j^T \vu_i$ and $\mC_j \vv_i$ is undesirable because pCCA is not being forced to identify new canonical weights, and therefore new correlations, across iterations.
We experimentally observed a highly collinear behaviour when using the Hotelling deflation scheme in our previous work on employing SCCA to BRCA~\citep{subramanian2018correlating}. 

This highlights the need for a better mechanism to generate multiple directions for SCCA and other pCCA variants. 
In our recent work, we presented a matrix deflation scheme for generating higher dimensional embeddings with pCCA variants~\citep{subramanian2021multimodal}.  This update scheme can be shown to reduce to a normalized variant of the Hotelling update scheme as shown in Appendix B, and runs into the same issue. In this work, we only compare our novel deflation schemes with the original Hotelling deflation scheme presented in~\eqref{HotellingDeflationSCCA}.

The key modification we make in the Hotelling update scheme (Figure~\ref{fig:hd}) is to update the data matrices $\mX_j$, $\mY_j$ instead of updating the cross-correlation matrix $\mC_j = \mX_j \mY_j^T$ at iteration $j$. Directly updating the data matrices removes all contributions of the found canonical weights, enforcing orthogonality of canonical weights across iterations and eliminating the repetition of correlations across iterations. We propose two approaches for this - the projected deflation and orthogonalized projected deflation (Figures~\ref{fig:pd}-\ref{fig:opd}). These deflation schemes are adapted from the matrix deflation schemes for sparse PCA~\citep{mackey2009deflation}. 

In the $j^{th}$ iteration of pCCA with projected deflation (PD)~(Figure~\ref{fig:pd}), the steps are
\begin{align*}
(\vu_j, \vv_j) &=  \text{ pCCA}(\mX_{j-1}, \mY_{j-1}), \nonumber \\
\mX_j &= (\mI_p - \vu_j \vu_j^T) \mX_{j-1}, \nonumber \\
\mY_j &= (\mI_q - \vv_j \vv_j^T) \mY_{j-1},
\end{align*}
where the initial values of $\mX_0$ and $\mY_0$ are set to the data matrix $\mX$ and $\mY$ respectively, 
and $\vu_j$, $\vv_j$ are assumed to be normalized to unit norm. 
At each deflation iteration, the projected deflation scheme projects the data matrices $\mX_{j-1}$ and $\mY_{j-1}$ on the space orthogonal to the newly found vector $\vu_j$ and $\vv_j$ respectively. Intuitively, this implies that the next iteration of pCCA will find no correlation by selecting the vectors $\vu_{j}$ and $\vv_{j}$ again at iteration $j+1$. Mathematically, if $\vu_j^T \vu_j = 1$,
\begin{align*}
\mC_j^T \vu_j &= \mY_j   \mX_j^T \vu_j  = \mY_j \mX_{j-1}^T(\mI_p - \vu_j \vu_j^T)  \vu_j  \nonumber \\
			  &= \mY_j \mX_{j-1}^T (\vu_j - \vu_j \vu_j^T \vu_j)  = 0, 
\end{align*}
and similarly $\mC_j \vv_j = 0$.
Thus, PD ensures that $\mC_j^T  \vu_j=0, \mC_j \vv_j = 0$. 
Therefore, the value $\vu^T \mC_j \vv$ obtained by picking $\vu$ as $\vu_j$ at the next iteration, or picking $\vv$ as $\vv_{j}$ in the next iteration, will be $0$.
However, there is no constraint enforced on how future vectors $\vu_i, i>j$ relate to $\vu_j$ (and similarly for $\vv_j$). 
To further improve the deflation scheme, we can look at the components of newly found vectors $\vu_{j}$ and $\vv_j$ which are in the space orthogonal to all previous vectors. That is, we can perform \textit{orthogonalized projected deflation}~(OPD) as  
\begin{align}
(\vu_j, \vv_j) &=  \text{pCCA}(\mX_{j-1}, \mY_{j-1}), \nonumber \\
\vr_j &= {\Tilde{\vr}_j}/{\norm{\Tilde{\vr}_j}}, \quad \Tilde{\vr}_j = (\mI_p - \mR_{j-1}\mR_{j-1}^T)\vu_j, \label{r_update}  \\
 \vs_j &= {\Tilde{\vs}_j}/{\norm{\Tilde{\vs}_j}}, \quad \Tilde{\vs}_j = {(\mI_q - \mS_{j-1}\mS_{j-1}^T)\vv_j}, \label{s_update}  \\ 
\mX_j &= (\mI_p - \vr_j \vr_j^T) \mX_{j-1}, \nonumber  \\
\mY_j &= (\mI_q - \vs_j \vs_j^T) \mY_{j-1}, \nonumber
\end{align}
for $j\geq 1$, where $\vr_1 = \vu_1$, $\vs_1 = \vv_1$, $\mR_{j-1} = [\vr_1 \dots \vr_{j-1}]$, and $\mS_{j-1} = [\vs_1 \dots \vs_{j-1}]$.  Note that (i) $\vr_1 \dots \vr_{j-1}$ are orthonormal vectors (ii) $\vr_1 \dots \vr_{j-1}$ form the basis of the space $\calR_{j-1}$ spanned by vectors $\vu_1 \dots \vu_{j-1}$, (iii) $\mR_{j-1}\mR_{j-1}^T$ is the projection matrix onto the space $\calR_{j-1}$. Similarly, $\mS_{j-1}\mS_{j-1}^T$ is the projection matrix onto the space spanned by vectors $\vv_1 \dots \vv_{j-1}$. By adding a memory element through matrices $\mR_{j-1}$ and $\mS_{j-1}$, OPD enforces additional orthogonality, as demonstrated in Figure~\ref{fig:opd}. 
By construction, $\mX_j$ lies in the space orthogonal to $\calR_{j}$ and $\vu_j \in \calR_j$, by~\eqref{r_update}. Therefore, $\vu_j^T \mX_j = 0 \ \forall j$. Similarly, $\forall i \> j$, $\mX_i$ lies in the space orthogonal to $\calR_i$, while $\vu_j \in \calR_j \subseteq \calR_i$. Extending similar statements to $\vv_j$ with respect to $\mY_i, i\geq j$, the OPD scheme exhibits
\begin{align*}
    \vu_j^T \mX_i &= 0, \ \forall i \geq j, \\
    \vv_j^T \mY_i &= 0, \ \forall i \geq j.
\end{align*}
This guarantees that the new canonical weights identified in later iterations of pCCA lie in a space orthogonal to already found canonical weights. This ensures that the new correlations are identified in newer subspaces, enabling more and diverse correlations to be captured. 

The proposed scheme of deflating the data matrices can be used for CCA, and pCCA with different structural penalties.
Table~\ref{deflation_cca_pcca} summarizes the different properties: no repetition of canonical weight pairs in consecutive iterations (P1), no repetition of canonical weights in consecutive iterations (P2), and no repetition of canonical weights across all iterations (P3). While all deflation schemes display the desired properties for CCA, the schemes differ for pCCA. 

Our proposed model for latent variable prediction makes use of pCCA and the proposed deflation schemes PD/OPD followed by a supervised prediction module~(Figure~\ref{fig:overview}). For each setting, the embeddings are generated in an iterative manner using deflation, as illustrated in~(Figure~\ref{fig:deflation_overview}). Among pCCA variants, we focus on SCCA and GN-SCCA. 
\begin{table}[]
\centering
\caption{Comparison of different deflation schemes for CCA and pCCA: Hotelling's deflation (HD), projected deflation (PD), orthogonalized projected deflation (OPD) for three properties -- P1 (no repetition of canonical weight pairs in consecutive iterations):  $\vu_j^T \mX_j \mY_j^T \vv_j = 0$, P2 (no repetition of canonical weights in consecutive iterations): $\vu_j^T \mX_j = \vv_j^T \mY_j = 0$, and P3 (no repetition of canonical weights across all iterations): $\vu_j^T \mX_i = \vv_j^T \mY_i = 0 \ \forall i > j$. }
\label{deflation_cca_pcca}
\resizebox{0.45\textwidth}{!}{%
\begin{tabular}{|c|c|c|c|c|c|c|}
\hline
Method & \multicolumn{2}{c|}{P1} & \multicolumn{2}{c|}{P2} & \multicolumn{2}{c|}{P3} \\
\hline
& CCA & pCCA & CCA  & pCCA & CCA & pCCA \\
\hline
HD & $\checkmark$ & $\checkmark$ & $\checkmark$  & -  & $\checkmark$ & -  \\
PD & $\checkmark$ & $\checkmark$ & $\checkmark$ & $\checkmark$ & $\checkmark$ & - \\
OPD & $\checkmark$ & $\checkmark$ & $\checkmark$ & $\checkmark$ & $\checkmark$ & $\checkmark$ \\
\hline
\end{tabular}
}
\end{table}

\section{Experiments and Results}\label{sec:experiments}

In this section, we systematically evaluate our proposed prediction pipeline. We begin by working with simulated data to demonstrate the potential of the pCCA-based embeddings in latent variable prediction, and the deflation schemes. 
We then demonstrate the use of our proposed methods on the TCGA-BRCA data for survival prediction. 

\subsection{Simulations}

\subsubsection{Simulated data}
We simulate data according to the model in Theorem~\ref{theorem:prob_cca}. 
For $i \in \{1 \dots N\}$, we first sample $\vz_i \sim \calN(0, \mI_d)$, the hidden latent variable. We then sample the observation samples $\vx_i, \vy_i$ according to $\vx_i|\vz_i \sim \calN(\mW_x \vz, \sigma_p^2 \mI_p)$, $\vy_i|\vz_i \sim \calN(\mW_y \vz, \sigma_q^2 \mI_q)$. Here, the weight matrices $\mW_x \in \R^{p \times d} $ and $\mW_y \in \R^{q \times d}$ are chosen to represent two different structural settings:

\begin{enumerate}[(i)]
    \item The first setting, \textit{Sparse}, takes into account sparsity when generating the weight matrices $\mW_x$ and $\mW_y$. In particular, each row of the weight matrices $\mW_x$ and $\mW_y$ have a fraction $s$ non-zero entries, sampled from the normal distribution $\calN(0, 1)$. 
    \item The second setting, \textit{Graph}, assumes a graph structure of the weight matrices $\mW_x$ and $\mW_y$. Here, a connected graph is generated randomly for each modality. Each row of the corresponding weight matrix is generated iteratively as follows. The eigenvectors corresponding to the lowest $k$ non-zero eigenvalues of the graph are linearly combined using uniformly random weights in $[0, 1]$, and then projected onto the space orthogonal to the previous rows.  
\end{enumerate}

In our simulation experiments, we generate 10 different folds with $n=100, p=q=200, d=5$, $k = 5, \sigma_x=\sigma_y =0.1$, $s  = 0.25$. Each fold is split into 60-10-30\% training-validation-testing sets. 
We evaluate the different pCCA methods and deflation schemes on this simulated data. %
For the graph-based CCA methods, the underlying graphs are constructed using the empirical covariance matrices as done by~\cite{du2015gn}. For all pCCA methods, the hyper-parameter tuning is done on the validation set based on the sum of additional correlations across dimensions.

\subsubsection{Results}

For this simulated data, we evaluate the different methods on three metrics:
\begin{enumerate}
    \item the mean of {additional correlations} across iterations,  
    \item the extent of orthogonality of new variates given by  $(\norm{\mC_k^T \vu_j} + \norm{\mC_k \vv_j})/2$, and 
    \item the mean-squared error (MSE) in the estimation of the latent variable $\vz$ 
\end{enumerate}

\begin{table}[tb]
\centering
\caption{Mean of additional correlations identified across $d=5$ deflation iterations on simulated data ($n=100, p=200,  q=200, \sigma=0.1$) with sparse structure (top) and graph structure (bottom). Results summarized across 10 folds and shown in percentages \%. Higher values desired.}
\label{table:simulation_correlations}
\resizebox{0.485\textwidth}{!}{%
\begin{tabular}{|c|l|l|ll|}
\hline 
 & pCCA & HD & PD & OPD \\
\hline
\multirow{2}{*}{Sparse} & SCCA & 89.48 $\pm$ 8.64 & 99.57 $\pm$ 0.28 & \textbf{99.61} $\pm$ \textbf{0.36} \\
 & GN-SCCA & 92.69 $\pm$ 3.34 & 99.19 $\pm$ 0.31 & \textbf{99.21} $\pm$ \textbf{0.34} \\
\hline
\multirow{2}{*}{Graph} &  SCCA & 93.41 $\pm$ 3.22 & 94.83 $\pm$ 1.91 & \textbf{94.87} $\pm$ \textbf{1.00} \\
 & GN-SCCA & 81.13 $\pm$ 5.70 & 93.89 $\pm$ 0.93 & \textbf{94.32} $\pm$ \textbf{1.08} \\
\hline
\end{tabular}
}
\end{table}

\begin{figure}[htb]
    \centering
    \includegraphics[width=0.49\textwidth]{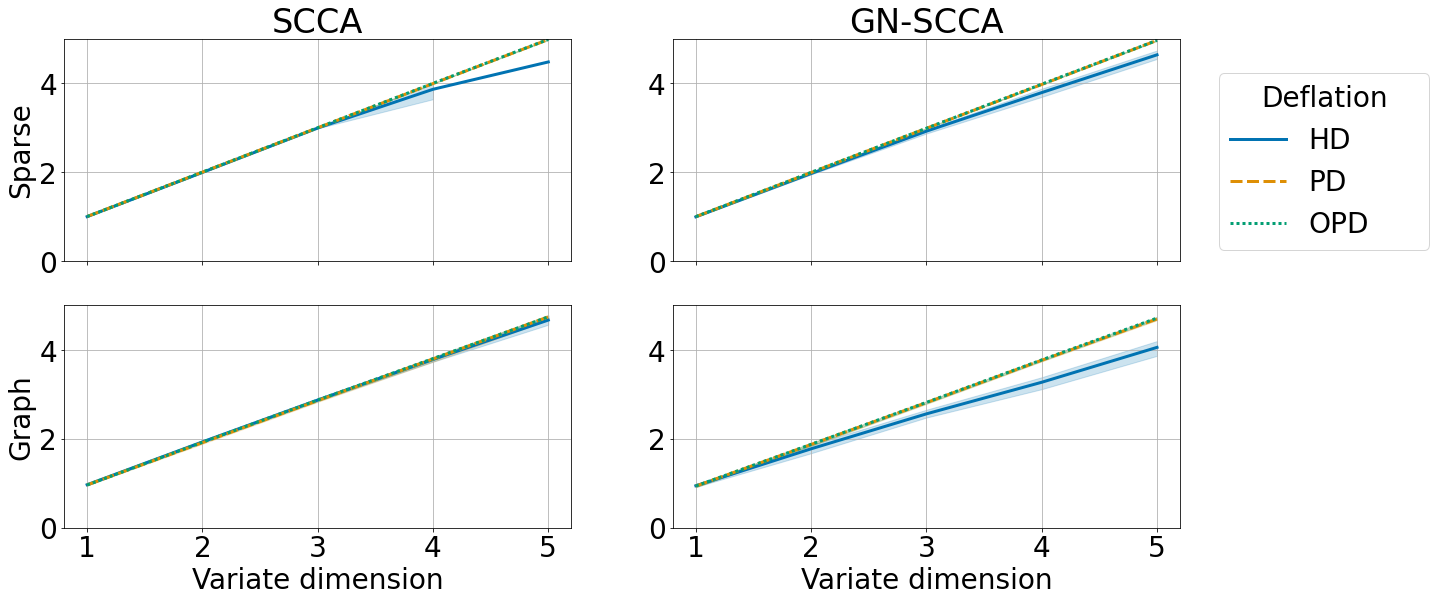}
    \caption{A plot of the sum of additional correlations identified across iterations  $\sum_{j=1}^k \tilde{\rho}_j$ summarized across 10 folds of simulations, with $k$ along the x-axis and the sum along the y-axis for sparse data (top) and graph-structured data (bottom) with SCCA and GN-SCCA. Higher values desired.}
    \label{fig:additional_correlation_simulation}
\end{figure}

\begin{figure}[htb]
    \centering
    \begin{subfigure}[c]{0.24\textwidth}
    \includegraphics[width=\textwidth,trim={1cm 5cm 0cm 5cm},clip]{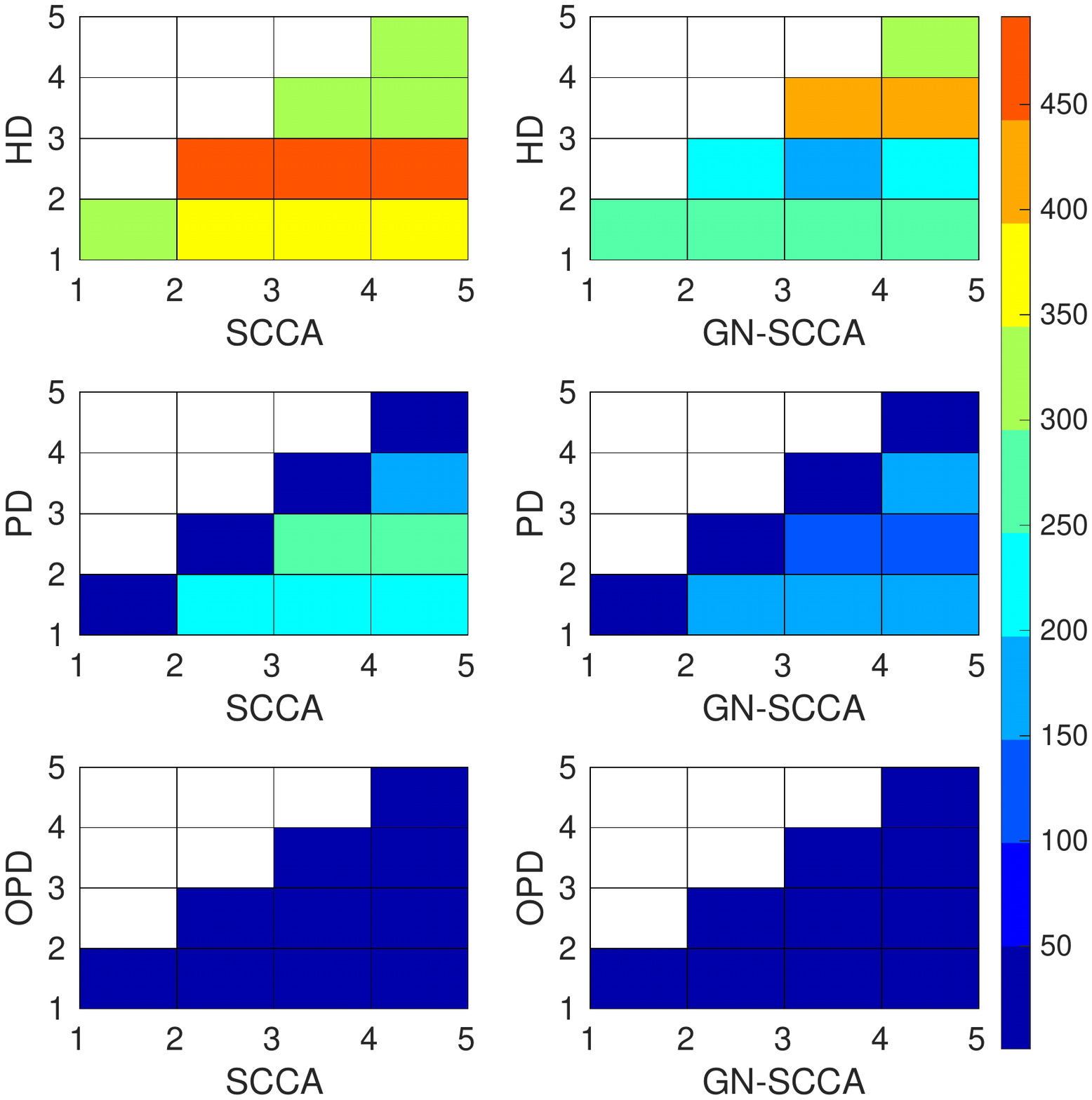}
    \caption{Graph data}
    \end{subfigure}
    \begin{subfigure}[c]{0.24\textwidth}
    \includegraphics[width=\textwidth,trim={1cm 5cm 0cm 5cm},clip]{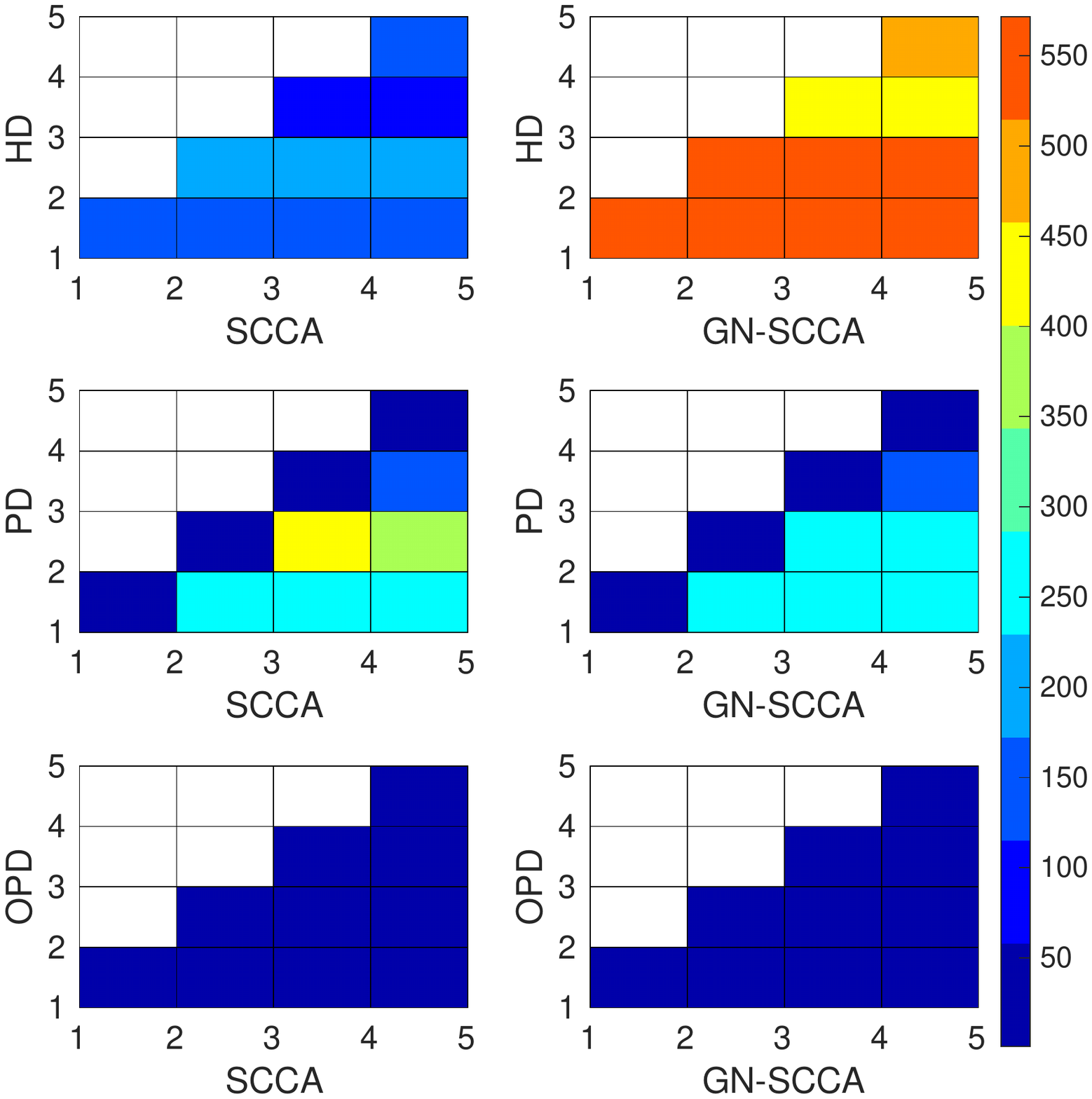}
    \caption{Sparse data}
    \end{subfigure}    
    \caption{Heatmap visualization of the orthogonality term $(\norm{\mC_k^T \vu_j} + \norm{\mC_k \vv_j})/2$ for $k \geq j$ with $k$ along the x-axis, and $j$ along the y-axis for different deflations across rows 
    for SCCA and GN-SCCA schemes. Results for graph-structured data (left) and sparse data (right). Lower values desired.}
    \label{fig:orthogonality_simulation}
\end{figure}

First, we compute the amount of additional (or new) correlation discovered across iterations which also provides an estimate of the new information contribution across covariate pairs. Specifically, let $\mR_{k-1}$ and $\mS_{k-1}$ denote the basis of the space covered by already found covariate vector pairs. Then, the additional correlation $\tilde{\rho}_k$ at each step $k$ is given by the correlation coefficient between $\vr_k^T \mX $ and $\vs_k^T \mY$ 
where $\mX$ and $\mY$ are the input data matrices and the vectors $\vr_k$ and $\vs_k$ capture the component of the covariate pair $(\vu_k, \vv_k)$ orthogonal to the space spanned by $\mR_{k-1}$ and $\mS_{k-1}$ respectively using
\begin{align*}
    \vr_k &= (\mI - \mR_{k-1} \mR_{k-1}^T) \vu_k, &
     \vs_k &= (\mI - \mS_{k-1} \mS_{k-1}^T) \vv_k. 
\end{align*}
The vectors $\vr_k$ and $\vs_k$ are appended as columns to the matrices $\mR_{k-1}$, $\mS_{k-1}$ to yield $\mR_k, \mS_k$. Under the definition above, the mean of additional correlations returned across the $d$ embedding dimensions $\frac{1}{d}\sum_{k=1}^d \tilde{\rho}_k$ of the different pCCA methods and deflation schemes are shown in Table~\ref{table:simulation_correlations}. The variation of sum of the new correlations across iterations is shown in Figure~\ref{fig:additional_correlation_simulation}. 
From these, we can observe that both our proposed deflation schemes can identify novel correlations better across simulation structures, especially with GN-SCCA.

Next, we measure the average extent of orthogonality across the two modalities for the pCCA problem with deflation schemes using the term ${\norm{\mC_k^T \vu_j}_2 + \norm{\mC_k \vv_j}_2}{/2}$
for $k \geq j$. A lower value indicates that the deflation scheme is encouraging diversity in the canonical variates, by removing existing components efficiently.
Heatmap visualizations of this is provided in Figure~\ref{fig:orthogonality_simulation} for one of the folds for each type of simulation structure. It can be observed, as expected, that the Hotelling deflation (HD) scheme does not enforce any of the variates to be orthogonal, the projected deflation (PD) scheme enforces orthogonality along the diagonal (low values across the diagonal), while the orthogonalized projected deflation (OPD) scheme enforces orthogonality across variates (low values throughout). 

\begin{table}[thb]
\centering
\caption{MSE in the prediction of the latent variable $\vz$ using the pCCA embeddings with an MLP compared to directly feeding observed data to an MLP (first row) on simulated data ($n=100, p=200,  q=200, \sigma=0.1$) for sparse and graph-structured simulation data. Lower values desired. }
\label{table:mean_squaredrror_simulation}
\begin{subtable}{0.5\textwidth}
\centering
\caption{Sparse data}
\resizebox{\textwidth}{!}{%
\begin{tabular}{|l|l|cc|c|}
\hline
 & Method & Modality 1 & Modality 2 & Concatenated \\ 
\hline
 & Original & 40.72 $\pm$ 6.45 & \textbf{39.74} $\pm$ \textbf{6.66} & 42.76 $\pm$ 8.73 \\
\hline
\multirow{2}{*}{HD} & SCCA & \textbf{35.25} $\pm$ \textbf{18.18} & 36.49 $\pm$ 16.68 & 36.40 $\pm$ 25.12 \\
 & GN-SCCA & 22.93 $\pm$ 18.44 & 24.72 $\pm$ 19.00 & \textbf{16.05} $\pm$ \textbf{11.03} \\
\hline
\multirow{2}{*}{PD} & SCCA & 29.04 $\pm$ 18.35 & 29.32 $\pm$ 19.88 & \textbf{24.17} $\pm$ \textbf{19.21} \\
 & GN-SCCA & 20.57 $\pm$ 9.14 & 20.38 $\pm$ 7.71 & \textbf{16.63} $\pm$ \textbf{6.21}\\
\hline
\multirow{2}{*}{OPD} & SCCA & \textbf{19.74} $\pm$ \textbf{13.66} & 21.28 $\pm$ 15.05 & 20.26 $\pm$ 15.69 \\
 & GN-SCCA & 23.40 $\pm$ 9.58 & 22.58 $\pm$ 8.65 & \textbf{17.42} $\pm$ \textbf{7.69} \\
\hline
\end{tabular} 
}
\end{subtable}
\newline
\vspace*{0.25cm}
\newline
\begin{subtable}{0.5\textwidth}
\centering
\caption{Graph-structured data}
\resizebox{\textwidth}{!}{%
\begin{tabular}{|l|l|cc|c|}
\hline
 & Method & Modality 1 & Modality 2 & Concatenated \\
\hline
 & Original & \textbf{74.28} $\pm$ \textbf{13.03} & 75.30 $\pm$ 6.51 & 75.50 $\pm$ 8.37 \\
\hline
\multirow{2}{*}{HD} & SCCA & 21.24 $\pm$ 13.82 & 21.98 $\pm$ 12.53 & \textbf{18.66} $\pm$ \textbf{9.42} \\
 & GN-SCCA & 34.30 $\pm$ 12.27 & 27.95 $\pm$ 10.59 & \textbf{20.74} $\pm$ \textbf{8.04} \\
\hline
\multirow{2}{*}{PD} & SCCA & 15.34 $\pm$ 8.69 & 14.51 $\pm$ 7.68 & \textbf{12.90} $\pm$ \textbf{7.03} \\
 & GN-SCCA & 31.23 $\pm$ 6.61 & 29.64 $\pm$ 8.13 & \textbf{26.06} $\pm$ \textbf{9.57} \\
\hline
\multirow{2}{*}{OPD} & SCCA  & 14.50 $\pm$ 8.10 & 14.61 $\pm$ 6.36 & \textbf{12.16} $\pm$ \textbf{6.63} \\
 & GN-SCCA & 27.64 $\pm$ 7.12 & 27.63 $\pm$ 9.84 & \textbf{24.04} $\pm$ \textbf{10.29} \\
\hline
\end{tabular} }
\end{subtable}
\end{table} 

Lastly, we evaluate the predictive potential of the resulting pCCA-based embeddings from the different pCCA methods using the different deflation schemes. 
To be consistent across the different variants of pCCA, we learn the latent variable prediction using the machine learning model of multi-layer perceptron (MLP). Specifically, we feed in the concatenated embedding vector $\begin{bmatrix}
\mU^T \mX \\ \mV^T \mY \end{bmatrix}$ to an MLP regressor from \texttt{scikit-learn}. As baselines, we also feed the original data (Modality 1 and Modality 2), the concatenated data (Concatenated) and the single-modality embeddings from the different methods as inputs to MLPs. When dealing with single-modality, we use an MLP with 50 hidden neurons, and 100 hidden neurons for the concatenated inputs. 
The mean-squared errors between the ground truth $\vz$ and the predicted $\hat{\vz}$ is summarized across folds in Table~\ref{table:mean_squaredrror_simulation} for the two structural settings of simulations. It is observed that although concatenation of the original data inputs confuses the MLP predictor, the use of pCCA benefits from the concatenated inputs across simulation settings. 
Further the proposed deflation schemes PD and OPD improve performance over HD, especially with SCCA. 

\subsection{Breast cancer data}\label{subsec:data}
Having demonstrated the potential of our proposed method on simulated data, we now proceed to work with real-world cancer data. 
In the context of breast cancer multi-modality data, the underlying survival state (or risk of death) is evident in both genomics and imaging features. The particular genomic signature of the cancer determines the aggressiveness of the cancer while the imaging features provides information about the local intensity and impact of the cancer. Hence, it is reasonable to expect, the probabilistic CCA model also applies to the breast cancer survival prediction. 
Therefore, using the two modalities jointly with the pCCA framework will provide a better estimate of survival, than working with single modalities independently. 

In this subsection, we first introduce the TCGA breast adenocarcinoma (BRCA) data and discuss the feature extraction pipeline before presenting the experimental results. 


\begin{figure}[thb!]
\centering
\begin{subfigure}[c]{0.15\textwidth}
\includegraphics[width=\textwidth]{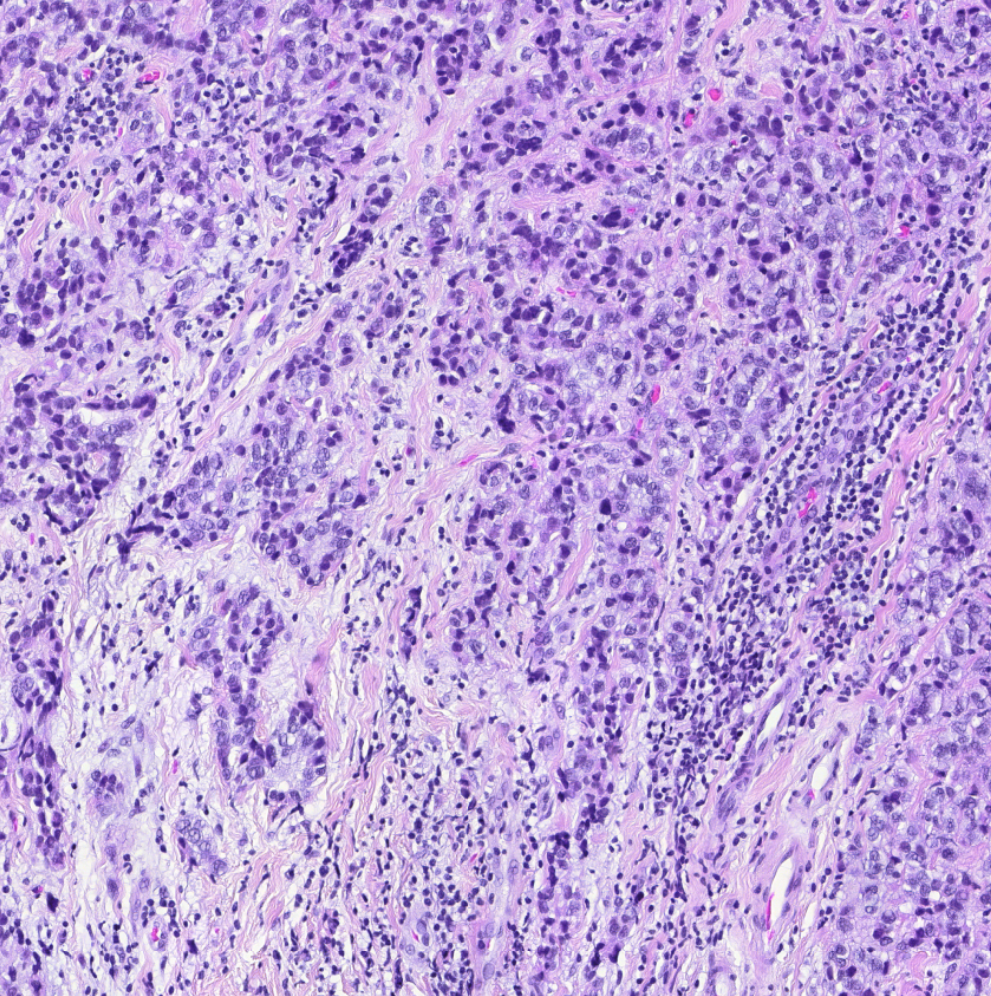}
\caption{}
\end{subfigure}
\begin{subfigure}[c]{0.15\textwidth}
\includegraphics[width=\textwidth]{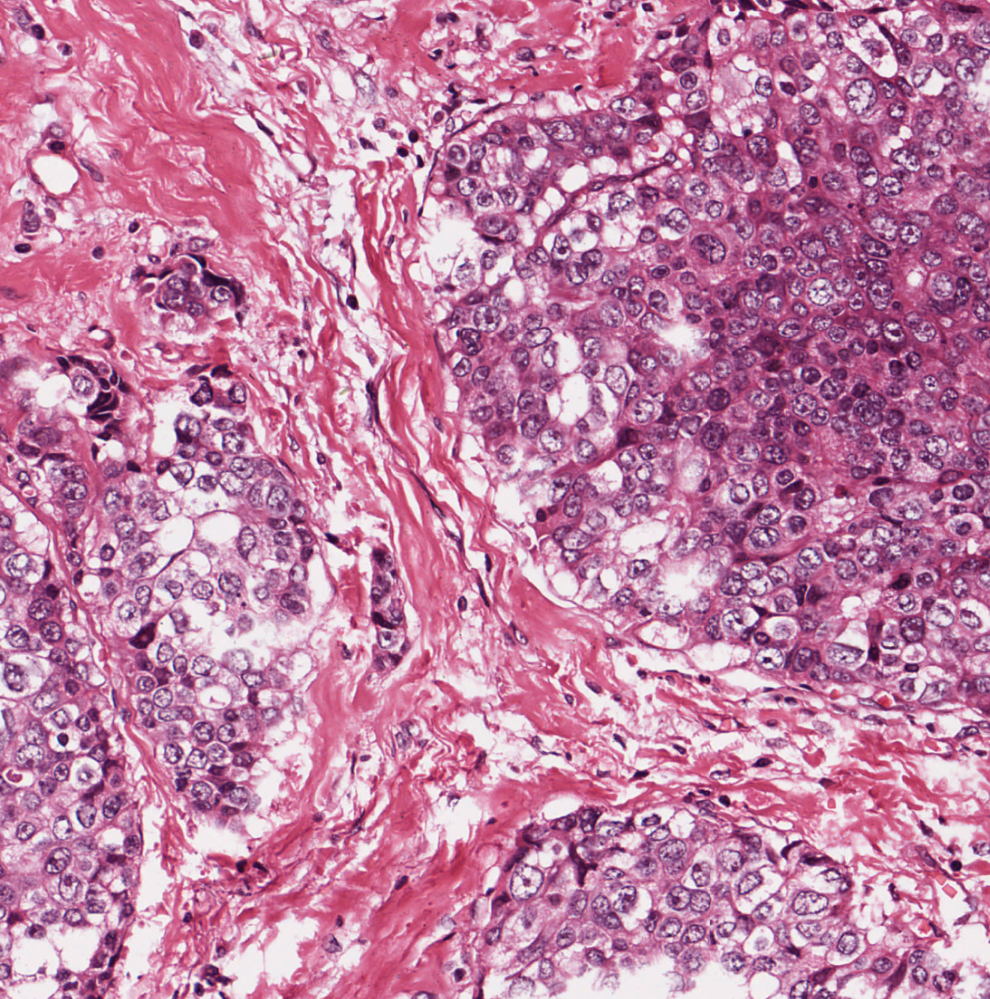}
\caption{}
\end{subfigure}
\begin{subfigure}[c]{0.15\textwidth}
\includegraphics[width=\textwidth]{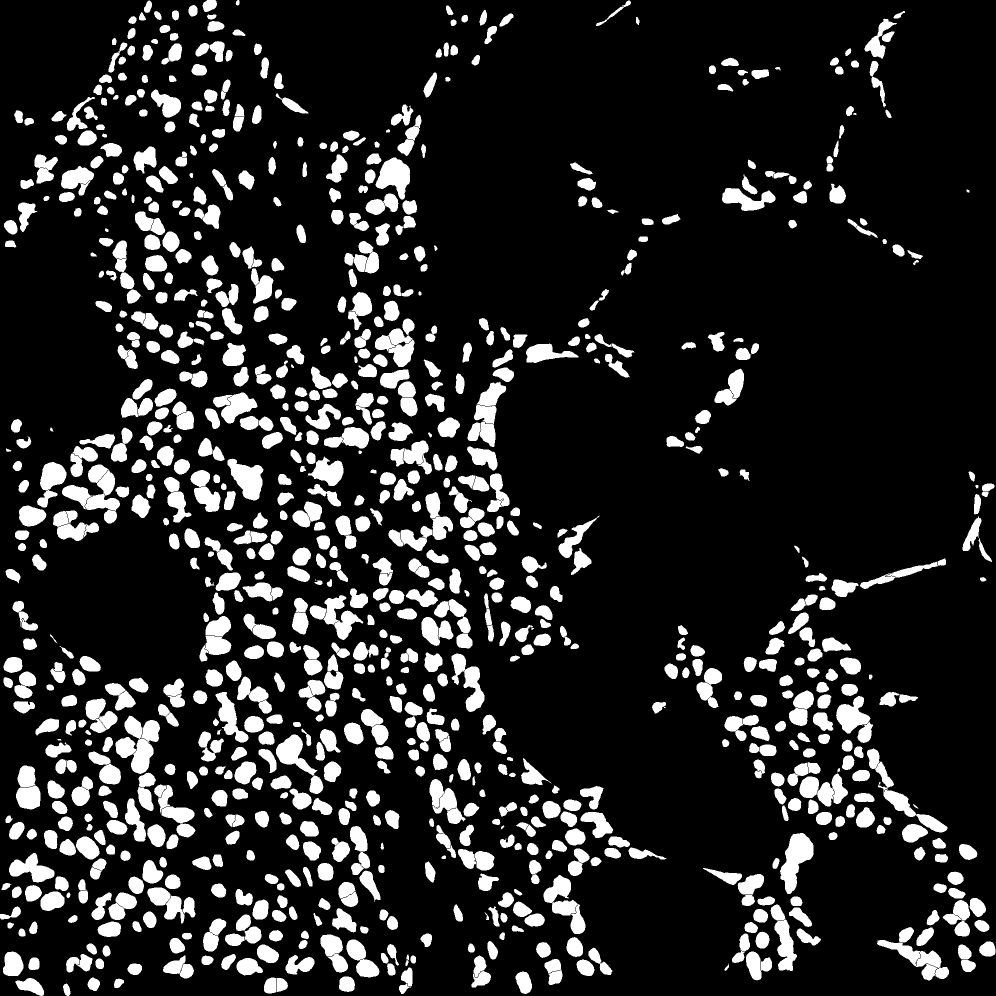}
\caption{}
\end{subfigure}
\caption{(a-b)~Examples of histology images from TCGA-BRCA , and (c)~an example of segmentations output by~\cite{hou2019robust}. Best viewed in color. }
\label{fig:tcga-brca-data}
\end{figure}

\subsubsection{Data, preprocessing, and prediction model}

We work on histology imaging and RNA-sequencing (RNA-seq) expressions from the TCGA-BRCA dataset of $n=974$ patients to demonstrate the potential on real data. For our experiments, this data is split into 60-15-25\% training-validation-testing sets. 

The histology imaging data was acquired from the National Cancer Institute's
\href{https://portal.gdc.cancer.gov/}{Genomic Data Commons portal}. Corresponding to each patient, we have the histology slides of over 20,000 x 20,000 pixels in size. We downloaded the nuclei segmentations corresponding to the histology images from a recently published adversarial learning framework~\cite{hou2019robust}. These were converted in 2000 x 2000 pixel patches to generate nuclei segmentation masks across all patches for all patients.  Example images and nuclei segmentation are shown in Figure~\ref{fig:tcga-brca-data}. 

To extract imaging features in a computationally feasible manner, we randomly selected $25$ patches of size 2000 x 2000 pixels for each patient. We fed these histology patches and the corresponding segmentation masks to the CellProfiler tool~\citep{carpenter2006cellprofiler} to extract area, shape and texture properties for each nuclei and cell in the patch - examples of which include area, compactness, eccentricity, Euler number and Zernike moments. 
These features were summarized across different patches of the same patient, using a 5-bin histogram for each of the 215 extracted features, yielding $1075$-dimensional imaging feature vectors for each patient. The imaging feature can be expanded to higher dimensions capturing more and diverse features. In particular, techniques using deep learning and convolutional neural networks can be employed for feature extraction or for an end-to-end feature learning, for example if deep learning based CCA is employed instead of pCCA.

The RNA-seq expression data was downloaded from the \href{http://firebrowse.org/}{FireBrowse platform} for all the TCGA-BRCA patients. The gene expression is available for over 20,000 genes. For computational feasibility with the pCCA-based framework, a subset of genes need to be selected. To do so, we evaluated the most variant genes using the coefficient of variation (the ratio $\sigma/\mu$ of standard deviation $\sigma$ and the mean $\mu$) of the log2-transformed expression values. We selected the top $1000$ genes based on the coefficient of variation, and the corresponding z-scores of the genes served as the genomic feature vector for each patient. 

The resulting genomics and imaging features, of dimensions $1000$ and $1075$ respectively, are input to our proposed prediction pipeline to generate joint embeddings using pCCA and the deflation schemes. 
The computed embedding vectors from samples can be used for survival prediction using models like Cox proportional hazards model and random survival forests. 
Survival prediction trains models to correctly assign a risk-value to each sample. With survival data, it is not necessary that all samples have experienced the event of interest. 
We work with the Cox proportional hazards model which is a commonly used semi-parametric model in survival analysis. 
The elastic net regularization is frequently imposed on the weights of the model~\citep{friedman2010regularization}. 
We use the algorithm's implementation from the \texttt{lifelines} Python package. 

\subsubsection{Results}
The proposed method is run on the TCGA-BRCA data with the imaging and genomic features as described above. As with simulated data, we evaluate the amount of average additional correlations, the extent of orthogonality and the predictive performance of the different methods with different deflation schemes. 

\begin{figure}[htb]
    \centering
    \includegraphics[width=0.495\textwidth]{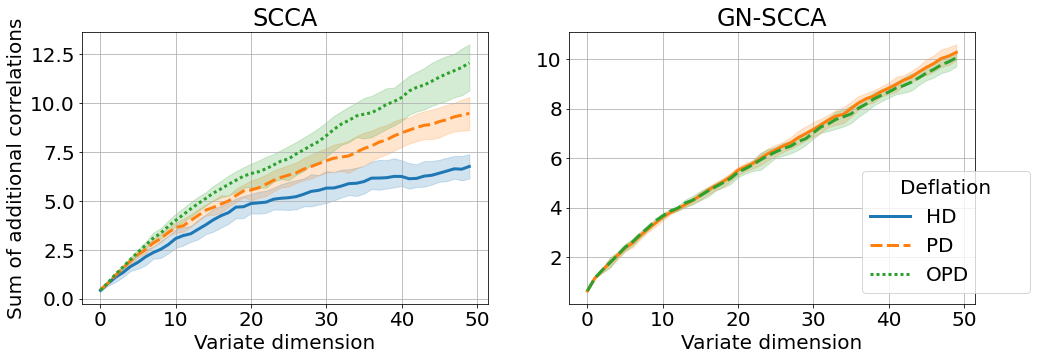}
    \caption{A plot of the sum of additional correlations identified across iterations  $\sum_{j=1}^k \tilde{\rho}_j$ summarized across 5 folds, with $k$ along the x-axis and the sum along the y-axis for the TCGA-BRCA data with SCCA and GN-SCCA. Higher values desired.}
    \label{fig:additional_correlation_brca}
\end{figure}

The average of the additional correlations across deflation iterations is reported in Table~\ref{tab:brca_corr_novelty}. The variation of the sume of additional correlations is plotted in Figure~\ref{fig:additional_correlation_brca}. When using real data, the graph-based pCCA does not converge with the HD scheme, while PD and OPD are both applicable. The PD and OPD schemes greatly improve the correlations returned by SCCA.

\begin{figure}
    \centering
    \includegraphics[width=0.475\textwidth,trim={1cm 1cm 0cm 0cm},clip]{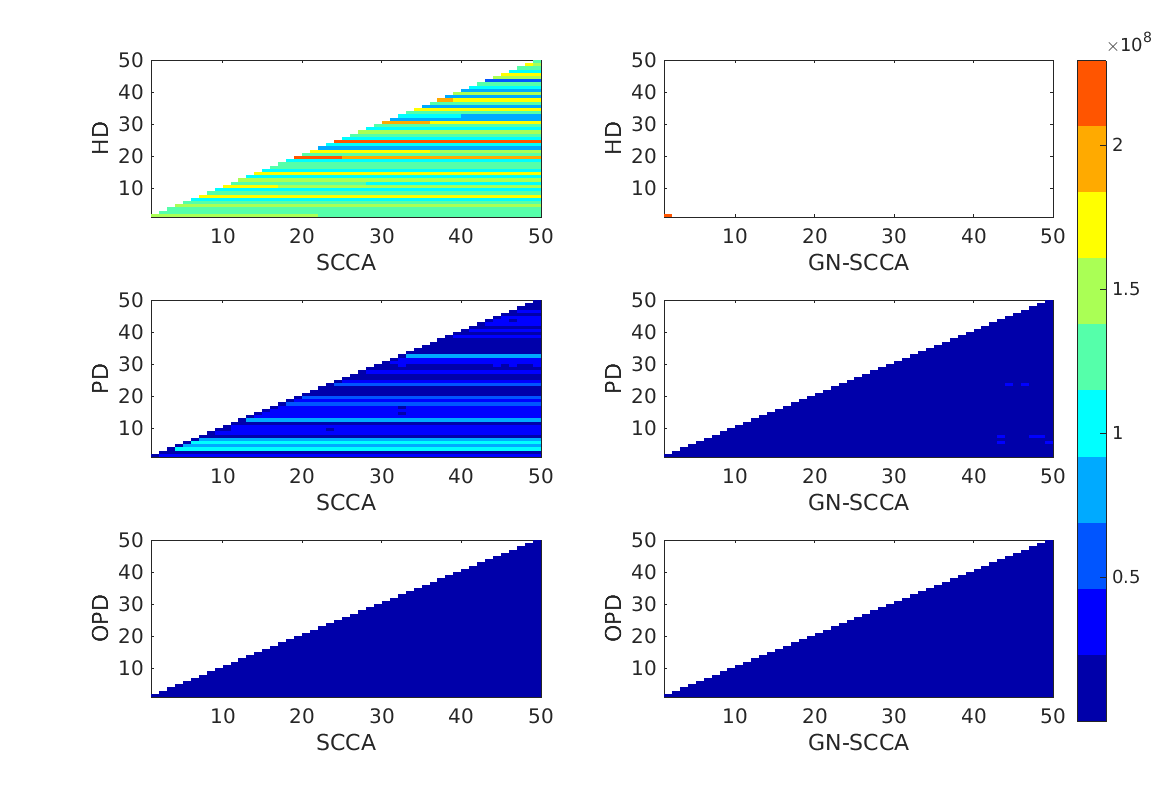}
    \caption{Heatmap visualization of the orthogonality term $(\norm{\mC_k^T \vu_j} + \norm{\mC_k \vv_j})/2$ for $k \geq j$ with $k$ along the x-axis, and $j$ along the y-axis for different deflation schemes across rows 
    for SCCA and GN-SCCA schemes. Results shown for one fold of the TCGA-BRCA data. Lower values desired.}
    \label{fig:new_variate_orthgonality_brca}
\end{figure}

The extent of orthogonality is visualized as before in Figure~\ref{fig:new_variate_orthgonality_brca} for one of the 5 folds. From the plot, it can be seen that our proposed PD and OPD schemes do better at encouraging orthogonality of the later variates for this fold. The same behaviour is also observed across folds.

\begin{table}[tb]
    \centering
    \caption{The average additional correlations across the 50 iterations of deflation used to generate the pCCA-based embeddings with TCGA-BRCA data. Higher values desired.}
    \label{tab:brca_corr_novelty}
\resizebox{0.425\textwidth}{!}{%
    \begin{tabular}{|l|c||cc|}
    \hline
    Method 		&HD		 & PD 	 & OPD \\
    \hline
SCCA	 & 6.76 $\pm$ 0.87	 & 9.47 $\pm$ 1.10	 & 12.05 $\pm$ 1.56 \\
GN-SCCA	 & -	 & 10.29 $\pm$ 0.43	 & 10.07 $\pm$ 0.42 \\
\hline
    \end{tabular}}
    
\end{table}

Lastly, we utilize the feature embeddings from different pCCA variants and different deflation methods for assigning risk of death in TCGA breast cancer patients using the Cox proportional hazards (CoxPH) model. Our performance evaluation is based on concordance indices (C-indices), which
quantifies the effectiveness of a given risk-prediction algorithm in correctly ordering events. Consider survival data $\{\ell_i = (e_i, t_i)\}_{i=1}^N$ indicating whether the patient died ($e_i = 1$) or not ($e_i = 0$) during the observation period, and the time $t_i$ at which the patient was last alive/observed.  
Let $\{o_i\}_{i=1}^N$ denote the predicted risk-score for all $N$ patients in the data. It is desired to predict higher risk scores $o_i$ for patients with a lower $t_i$ if the event was observed, $e_i = 1$. This is captured in the C-index, defined as
\begin{equation*}
\text{C-index} = \frac{1}{n} \sum_{i|  e_i = 1} \sum_{t_j > t_i} \mathbf{1}[o_i > o_j],
\end{equation*} where $n$ is the number of ordered pairs in the ground-truth data. 
High C-index values are desired. A C-index of 0.5 is equivalent to a random guess. 


For each fold, we fit the CoxPH model on the training set and report performance on the testing set. We impose the elastic net penalty with a penalizer factor of 0.1. The C-indices for the different methods are in Table~\ref{tab:survival_prediction}. We run the CoxPH model with different inputs for each setting. The first row reports the C-indices on a 100-dimensional PCA embedding generated from all genes, a 100-dimensional PCA embedding generated from the imaging feature, and their concatenation. Similarly, later rows report the C-indices of using embeddings from different pCCA and deflation schemes with genomics embeddings $\mU^T \mX$ only, imaging embedding $\mV^T \mY$ only, and the concatenation of both as inputs $\begin{bmatrix} \mU^T \mX \\ \mV^T \mY \end{bmatrix}$ to the CoxPH. 
In our experiments, GN-SCCA was unable to train successfully when using HD. However, GN-SCCA worked with PD and OPD schemes.

\begin{table}[!htb]
    \centering 
    \caption{The C-indices returned for survival prediction of TCGA-BRCA using CoxPH with different inputs. Results summarized across the 5-folds. Higher values desired.}
    \label{tab:survival_prediction}
\resizebox{0.485\textwidth}{!}{%
\begin{tabular}{|l|l|cc|c|}
\hline
& Method &	 Genomics &	 Imaging & 	Concatenated\\
\hline
 	&	PCA	 & 65.76 $\pm$ 9.02	 & 61.24 $\pm$ 6.39	 & \textbf{66.86} $\pm$ \textbf{10.40} \\
\hline
\multirow{2}{*}{HD}	&	SCCA	 & \textbf{65.95} $\pm$ \textbf{10.70}	 & 54.00 $\pm$ 11.23	 & 65.71 $\pm$ 7.96\\
&	GN-SCCA	 & - & - & -\\
\hline 
\multirow{2}{*}{PD}		&	SCCA	 & \textbf{68.34} $\pm$ \textbf{8.40}	 & 55.66 $\pm$ 11.14	 & 67.37 $\pm$ 6.58\\
	&	GN-SCCA	 & 64.94 $\pm$ 8.98	 & 56.81 $\pm$ 13.31	 & \textbf{66.42} $\pm$ \textbf{12.08}\\
\hline
\multirow{2}{*}{OPD}		&	SCCA	 & \textbf{67.63} $\pm$ \textbf{9.43}	 & 54.19 $\pm$ 11.23	 & 65.25 $\pm$ 5.61\\
	&	GN-SCCA	 & 65.52 $\pm$ 7.67	 & 57.87 $\pm$ 10.28	 & \textbf{68.08} $\pm$ \textbf{9.88}\\
	\hline
\end{tabular}
}
   
\end{table}

From the C-index performance, it can be observed that while the use of the existing HD deflation scheme with pCCA results in a poorer performance than PCA embeddings, the use of PD and OPD schemes has the potential to improve performance. This is evidenced in the concatenated GN-SCCA embeddings with PD and OPD. The concatenated GN-SCCA embeddings perform better than single modality GN-SCCA embeddings. 
The poorer performance of imaging features compared to genomics features across the table highlight the need to improve these features by adding more complexity in the histology feature extraction. 
 \section{Conclusion}\label{section:conclusions}

In this work we investigated the use of CCA-based embeddings for latent variable prediction. We first proved mathematically why the joint posterior mean estimators are better than combining individual modality posterior mean estimators for latent variable prediction using a probabilistic model of two-modality data. We recalled that CCA provides the maximum likelihood estimates of the model parameters for the studied graphical model. Using this result, we proposed a two-stage prediction model with CCA. Next, we proposed two deflation schemes that can help generate informative, multi-dimensional embeddings when working with the penalized versions of CCA, pCCA. 

We demonstrated the efficacy of our proposed two-stage prediction model on simulated data and real data from TCGA-BRCA histology and RNA-seq data. Our method improves latent variable prediction across different settings of simulated data, with desirable properties of embeddings. 
On the TCGA-BRCA data, we discovered better embeddings with desirable properties that displayed potential to improve survival prediction. 
Overall, our results highlight the importance of intelligently combining multi-modality data. 
In this work, by focusing on the shared information captured by pCCA variants, we are able to improve latent variable prediction performance. 

Our work faces a few limitations. Although the use of penalized variants is motivated by the possibility to use prior knowledge, we did not make use of any such information from protein-protein interaction networks, gene regulatory networks or molecular pathways in biology. Incorporating these prior knowledge could greatly benefit survival prediction. In addition, our method does not tackle non-linear or more complex embeddings, for example those generated from neural networks. It would be interesting to explore how multi-modality fusion can be performed using deep learning methods in a systematic way with theoretical backing. 


 
\setcounter{equation}{0}
\def\theequation{A\arabic{equation}}

\appendices




\appendices


\newpage
\section{Error in estimation of latent variables}\label{sec:reduction_in_variance}

We work with a two-modality data model as motivated in~\cite{bach2005probabilistic}.
Consider
\begin{align*}
    \vz &\sim \calN(\vzero, \mI_d), \\
    \vx  &= \mW_x \vz + \bm{\epsilon}_x, \bm{\epsilon}_x \sim \calN(\vzero, \sigma_x^2 \mI_p), \\
    \vy &= \mW_y \vz + \bm{\epsilon}_y, \bm{\epsilon}_y \sim \calN(\vzero, \sigma_y^2 \mI_q),
\end{align*}
where $\vz \in \R^d, \vx \in \R^p, \vy \in \R^q$, the combination matrices are $\mW_x \in \R^{p \times d}$ and $\mW_y \in \R^{q \times d}$ and random vectors $\vz, \bm\epsilon_x, \bm\epsilon_y$ are independent. Assume that the model parameters $\mW_x, \mW_y, \sigma_x$ and $\sigma_y$ are all known.
Let $\hat\vz_c= \E[\vz|(\vx, \vy)]$ be posterior mean estimator which estimates $\vz$ using the two modalities jointly and  $\hat\vz_\beta = \beta \E[\vz|\vx] + \Bar\beta \E[\vz|\vy]$ for $\beta \in [0,1], \Bar\beta = 1-\beta$ be any $\beta$ linear combination of the estimates based on single modalities $\vx$ and $\vy$.
\begin{thm}
\label{theorem_appendix}
Under the assumed data model with known model parameters $\mW_x, \mW_y, \sigma_x$ and $\sigma_y$, $\hat\vz_c$ is the better estimator compared to $\hat\vz_\beta$ in terms of $e(\hat\vz) = \E_{(\vx, \vy)|\vz}[\norm{\hat\vz - \vz}_2^2]$, the mean squared error in the estimation of $\vz$, with
\[e(\hat\vz_\beta) \geq e(\hat\vz_c) \quad \forall \beta \in [0,1]. \]
\end{thm}

If only a single modality $\vx$ is used for the estimation, we have the posterior mean estimator as
\begin{align*}
    \hat{\vz}_x &\triangleq \E [\vz|\vx] = (\mW_x \mW_x^T + \sigma_x^2 \mI)^{-1} \mW_x^T \vx \nonumber \\
    &= \underbrace{(\mW_x^T \Psi_x^{-1} \mW_x + \mI_d)^{-1} \mW_x^T \Psi_x^{-1}}_{\mG_x} \vx = \mG_x \vx, 
\end{align*}
where $\Psi_x = \sigma_x^2 \mI_p$.
Similarly, if only modality $\vy$ is used, the posterior mean estimator is 
\begin{align*}
    \hat{\vz}_y &\triangleq \E [\vz|\vy] = (\mW_y \mW_y^T + \sigma_y^2 \mI)^{-1} \mW_y^T \vy \nonumber \\
    &= \underbrace{(\mW_y^T \Psi_y^{-1} \mW_y + \mI_d)^{-1} \mW_y^T \Psi_y^{-1}}_{\mG_y} \vy = \mG_y \vy, 
\end{align*}
where $\Psi_y = \sigma_y^2 \mI_q$. 
A $\beta$-weighted linear mixing of these estimators is 
\begin{align*}
    \hat{\vz}_{\beta} &\triangleq \beta \hat{\vz}_x +  \Bar{\beta} \hat{\vz}_y = \beta \mG_x \vx + \Bar\beta \mG_y \vy \nonumber \\
    &= \underbrace{\begin{bmatrix}
    \mG_x & \mG_y 
    \end{bmatrix}\begin{bmatrix}
    \beta \mI_p & \vzero \\ \vzero & \Bar\beta \mI_q
    \end{bmatrix}}_{\mG_\beta} \begin{bmatrix}
    \vx \\ \vy
    \end{bmatrix} 
    = \mG_\beta \begin{bmatrix}
    \vx \\ \vy
    \end{bmatrix} ,
\end{align*}
where $\beta \in [0,1]$ and $\Bar{\beta} = 1 - \beta$. Note that this estimator $\hat{\vz_\beta}$ includes the estimators $\hat{\vz}_x$ and $\hat{\vz}_y$ as special cases for $\beta=1$ and $\beta=0$ respectively. 
Lastly, the modalities can be used jointly in the estimation of $\vz$ which leads to the estimator that combines $\vx$ and $\vy$ as
\begin{align*}
    \hat{\vz}_c &\triangleq \E[\vz | (\vx, \vy)] \nonumber \\
    &=  \underbrace{(\mW^T \Psi^{-1} \mW + \mI_d)^{-1} \mW^T \Psi^{-1}}_{\mG_c} \begin{bmatrix}
    \vx \\
    \vy
    \end{bmatrix} = \mG_c \begin{bmatrix}
    \vx \\ \vy
    \end{bmatrix},
\end{align*}
where $\mW = \begin{bmatrix}
\mW_x \\ \mW_y
\end{bmatrix}$ and $\Psi = \begin{bmatrix}
\Psi_x & \vzero \\ \vzero & \Psi_y
\end{bmatrix} =  \begin{bmatrix}
\sigma_x^2 \mI_p & \vzero \\ \vzero & \sigma_y^2 \mI_q
\end{bmatrix}$ such that $\Psi^{-1} = \begin{bmatrix}
\Psi_x^{-1} & \vzero \\ \vzero & \Psi_y^{-1}
\end{bmatrix} = \begin{bmatrix}
\frac{1}{\sigma_x^2} \mI_p & \vzero \\ \vzero & \frac{1}{\sigma_y^2} \mI_q
\end{bmatrix}$.


\begin{lemma}\label{lemma:error_expression}
For any estimator of the form $\hat\vz= \mG \begin{bmatrix}
\vx \\ \vy
\end{bmatrix}$ with $\mG \in \R^{d\times (p+q)}$, 
the error in the estimation of $\vz$ is
\begin{align*}
    e(\hat\vz) &= \vz^T (\mG \mW - \mI_d)^T (\mG \mW - \mI_d) \vz + \Trace{\mG \Psi \mG^T}.
\end{align*}
\end{lemma}

\begin{proof}
Expanding the error expression, we obtain
\begin{align}
    e(\hat{\vz}) 
    &= \E [(\hat{\vz} - \vz)^T(\hat{\vz} - \vz)] \nonumber \\
                       &= \E [( \mG \begin{bmatrix} \vx \\ \vy \end{bmatrix} - \vz)^T( \mG \begin{bmatrix} \vx \\ \vy \end{bmatrix} - \vz)] \nonumber \\
                       &= \E [( \mG \begin{bmatrix} \mW_x \vz + \bm\epsilon_x \\ \mW_y \vz + \bm\epsilon_y \end{bmatrix} - \vz)^T ( \mG \begin{bmatrix} \mW_x \vz + \bm\epsilon_x \\ \mW_y \vz + \bm\epsilon_y \end{bmatrix} - \vz) \nonumber \\
                    &= \E [( \mG \mW \vz + \mG \bm \epsilon- \vz)^T( \mG \mW \vz +\mG \bm\epsilon - \vz)] \nonumber\\
                       & = \E [\vz^T(\mG \mW - \mI_d)^T (\mG \mW - \mI_d) \vz ] +
                          \E[ \bm{\epsilon}^T \mG^T \mG \bm\epsilon] \nonumber \\ 
                      & = \vz^T(\mG \mW - \mI_d)^T (\mG \mW - \mI_d) \vz +
                          \E[ \bm{\epsilon}^T \mG^T \mG \bm\epsilon], \nonumber
\end{align}
where $\bm\epsilon = \begin{bmatrix}
\bm\epsilon_x \\ \bm\epsilon_y
\end{bmatrix}$, with $\bm \epsilon \sim \calN (\vzero, \Psi)$. Thus, $\mG \bm\epsilon \sim \calN (\vzero, \mG \Psi \mG^T)$, such that the second term of the error $e(\hat\vz)$ is
\begin{align*}
    \E[ \bm{\epsilon}^T \mG^T \mG \bm\epsilon] &= \E[ (\mG \bm\epsilon)^T \mG \bm\epsilon] \nonumber   \\
    &=  \E[ \Trace{ (\mG \bm\epsilon)^T \mG \bm\epsilon}]\\
    &=  \E[ \Trace{(\mG \bm\epsilon) (\mG \bm\epsilon)^T}] \nonumber \\
    &=  \E[ \Trace{\mG \Psi \mG^T}]\nonumber \\ 
    &=  \Trace{\mG \Psi \mG^T}, \nonumber
\end{align*}
where $\Trace{\mM}$ denotes the trace of matrix $\mM$. Therefore, 
\begin{align}
    e(\hat{\vz}) 
    & = \vz^T(\mG \mW - \mI_d)^T (\mG \mW - \mI_d) \vz  + \Trace{\mG \Psi \mG^T}. \nonumber 
\end{align}
\end{proof}

From Lemma~\ref{lemma:error_expression}, we obtain
\begin{align}
    e(\hat{\vz}_\beta) &=  \vz^T(\mG_\beta \mW - \mI_d)^T (\mG_\beta \mW - \mI_d) \vz  + \Trace{\mG_\beta \Psi \mG_\beta^T} \nonumber \\
    &=  \vz^T\mK_\beta^T \mK_\beta \vz  + \Trace{\mG_\beta \Psi \mG_\beta^T}, \ \text{and} \label{eq:e_beta}\\
    e(\hat{\vz}_c) &=  \vz^T(\mG_c \mW - \mI_d)^T (\mG_c \mW - \mI_d) \vz  + \Trace{\mG_c \Psi \mG_c^T}  \nonumber \\
    &=  \vz^T\mK_c^T \mK_c \vz  + \Trace{\mG_c \Psi \mG_c^T}, \label{eq:e_c} 
\end{align}
where $\mK_\beta = \mG_\beta \mW - \mI_d$ and $\mK_c = \mG_c \mW - \mI_d$.
\begin{lemma}\label{lemma:aboutK}
Let $\mK$ be given by $\mK = \mG \mW - \mI_d$ where $\mG~=~(\mW^T \Psi^{-1} \mW + \mI_d)^{-1} \mW^T \Psi^{-1} $ and $\Psi \succ 0$. Then, $\mK = -(\mW^T \Psi^{-1} \mW + \mI_d)^{-1}$ and is negative definite. 
\end{lemma}

\begin{proof}
Simplifying the expression for $\mK$, we obtain
\begin{align*}
\mK &= \mG \mW - \mI_d \nonumber \\
&= (\mW^T \Psi^{-1} \mW + \mI)^{-1} \mW^T \Psi^{-1}   \mW - \mI_d  \nonumber \\
&= (\mW^T \Psi^{-1} \mW + \mI)^{-1} (\mW^T \Psi^{-1}  \mW + \mI_d - \mI_d) - \mI_d  \nonumber \\
&= \mI_d -  (\mW^T \Psi^{-1} \mW + \mI)^{-1} -\mI_d \nonumber \\
&=  -  (\mW^T \Psi^{-1} \mW + \mI)^{-1}. 
\end{align*}
Further $\Psi \succ 0$,  $\implies \Psi^{-1} \succ 0 \implies \mW^T \Psi^{-1} \mW \succeq 0 \implies  \mW^T \Psi^{-1} \mW + \mI_d \succ 0 \implies  (\mW^T \Psi^{-1} \mW + \mI_d)^{-1} \succ 0$.
\end{proof}
\begin{cor}
$\mK_x = \mG_x \mW_x - \mI_d$,  $\mK_y = \mG_y \mW_y - \mI_d$, and  $\mK_c = \mG_c \mW - \mI_d$ are all negative definite. \end{cor}
\begin{proof}
For $\mK_x, \mK_y$, substitute $\mG, \mW$ with matrices $\mG_x, \mW_x$ and $\mG_y, \mW_y$  respectively. Setting $\mG = \mG_c$ yields the result for $\mK_c$. 
\end{proof}

\begin{cor}
$\mK_{\beta} = \mG_\beta \mW - \mI_d$ is negative definite.
\end{cor}
\begin{proof}
We observe that
\begin{align*}
 \mK_{\beta} &= \mG_\beta \mW - \mI_d    \\
     &= \begin{bmatrix}
        \mG_x & \mG_y 
        \end{bmatrix}\begin{bmatrix}
        \beta \mI_p & \vzero \\ \vzero & \Bar\beta \mI_q
        \end{bmatrix} \mW - \mI_d \\
    &= \beta (\mG_x \mW_x - \mI_d) + \Bar\beta (\mG_y \mW_y - \mI_d), \\
    &= \beta \mK_x + \Bar\beta \mK_y,
\end{align*}
where $\beta \in [0,1], \Bar\beta=1-\beta$. Thus, $\mK_\beta$ is negative definite.
\end{proof}

The matrices $\mK_x$ and $\mK_y$ can be decomposed using the Cholesky decomposition as 
\begin{align*}
    \mK_x &= - \mL_x \mL_x^T \\
    \mK_y &= - \mL_y \mL_y^T
\end{align*}
using lower triangular matrices $\mL_x, \mL_y \in \R^{d \times d}$ with positive diagonal entries,
such that 
\begin{align}
    \mK_\beta &= -\begin{bmatrix}
                \beta \mI_d & \vzero \\ \vzero & \Bar\beta \mI_d
                \end{bmatrix} \begin{bmatrix}
                \mL_x\mL_x^T \\ \mL_y \mL_y^T
                \end{bmatrix}   \nonumber \\
            &= -\begin{bmatrix}
                \mL_x & \mL_y
                \end{bmatrix}\begin{bmatrix}
                \beta \mI_d & \vzero \\ \vzero & \Bar\beta \mI_d 
                \end{bmatrix} \begin{bmatrix}
                \mL_x^T \\ \mL_y^T
                \end{bmatrix} \nonumber  \\ 
            &= -\mL_\beta \mL_\beta^T  \label{k_beta}
\end{align}
The matrix products $\mL_x\mL_x^T$ and $\mL_y\mL_y^T$ are positive definite, and thus invertible. 
For the joint estimator, the matrix $\mK_c$ is 
\begin{align}
    &\mK_c = \mG_c \mW - \mI_d  \nonumber \\
    &= (\mW^T \Psi^{-1} \mW + \mI_d)^{-1} \mW^T \Psi^{-1}\mW - \mI_d  \nonumber \\
    &= -(\mW^T \Psi^{-1} \mW + \mI_d)^{-1}  \nonumber \\
    &= -(\mW_x^T \Psi_x^{-1} \mW_x + \mW_y^T \Psi_y^{-1} \mW_y + \mI_d) ^{-1}  \nonumber \\
    &= -((\mL_x \mL_x^T)^{-1} + (\mL_y \mL_y^T)^{-1} - \mI_d) ^{-1} \nonumber \\
    &= - (\mL_y \mL_y^T) \Big( \underbrace{\mL_x \mL_x^T + \mL_y \mL_y^T - (\mL_x \mL_x^T) (\mL_y \mL_y^T)}_{\tilde{\mL}} \Big)^{-1} (\mL_x \mL_x^T) \nonumber \\
    &= - \begin{bmatrix}
        \mL_x & \mL_y
        \end{bmatrix} \begin{bmatrix}
        \vzero \\\mL_y^T
        \end{bmatrix} 
        \tilde{\mL}^{-1}  \begin{bmatrix}
        \mL_x  & \vzero 
        \end{bmatrix} \begin{bmatrix}
        \mL_x^T \\ \mL_y^T
        \end{bmatrix} \nonumber \\
    &= - \begin{bmatrix}
        \mL_x & \mL_y
        \end{bmatrix} \begin{bmatrix}
        \vzero & \vzero \\
        \mL_y^T\tilde{\mL}^{-1} \mL_x& \vzero 
        \end{bmatrix}
         \begin{bmatrix}
        \mL_x^T \\ \mL_y^T
        \end{bmatrix}.  \label{k_c}
\end{align}

\begin{lemma}
Let $\mK_\beta$ and $\mK_c$ be as defined in~\eqref{k_beta} and~\eqref{k_c}. Then, 
\[ \mK_\beta - \mK_c \preceq 0. \]
\end{lemma}
\begin{proof}
\begin{align*}
 \mK_\beta   - \mK_c &=  \begin{bmatrix}
\mL_x & \mL_y
\end{bmatrix}\begin{bmatrix}
-\beta \mI_d & \vzero \\ \vzero & -\Bar\beta \mI_d 
\end{bmatrix} \begin{bmatrix}
\mL_x^T \\ \mL_y^T
\end{bmatrix}  \nonumber  \\
& \qquad \qquad + \begin{bmatrix}
\mL_x & \mL_y
\end{bmatrix} \begin{bmatrix}
\vzero & \vzero \\
\mL_y^T\tilde{\mL}^{-1} \mL_x& \vzero 
\end{bmatrix} 
 \begin{bmatrix}
\mL_x^T \\ \mL_y^T
\end{bmatrix} \nonumber  \\
&=  \begin{bmatrix}
\mL_x & \mL_y
\end{bmatrix} \underbrace{\begin{bmatrix}
-\beta \mI_d  & \vzero \\
\mL_y^T\tilde{\mL}^{-1} \mL_x& -\Bar\beta \mI_d 
\end{bmatrix}  }_{\hat\mL}
 \begin{bmatrix}
\mL_x^T \\ \mL_y^T
\end{bmatrix}.
\end{align*}
 Since $\hat\mL$ is lower triangular with non-positive diagonal entries, $\hat\mL$ is negative semi-definite. Thus, $\mK_\beta - \mK_c$ is negative semi-definite. 
\end{proof}

Lastly, note that 
\begin{align*}
    \mG_\beta &= -\begin{bmatrix}
                \beta \mI_d & \vzero \\ \vzero & \Bar\beta \mI_d 
                \end{bmatrix} \begin{bmatrix}
                \mG_x \\ \mG_y
                \end{bmatrix} \nonumber \\
              &= -\begin{bmatrix}
                \beta \mI_d & \vzero \\ \vzero & \Bar\beta \mI_d 
                \end{bmatrix} \begin{bmatrix}
                \mK_x \mW_x^T \Psi_x^{-1} \\ \mK_y \mW_y^T \Psi_y^{-1} 
                \end{bmatrix}  \nonumber  \\
              &= -\underbrace{\begin{bmatrix}
                \beta \mI_d & \vzero \\ \vzero & \Bar\beta \mI_d 
                \end{bmatrix} \begin{bmatrix}
                \mK_x \\ \mK_y 
                \end{bmatrix}}_{\mK_\beta}   \mW^T \Psi^{-1}  \nonumber \\
              &= - \mK_\beta   \mW^T \Psi^{-1}, \ \text{and} \\
          \mG_c &= (\mW^T \Psi^{-1} \mW + \mI_d)^{-1} \mW^T \Psi^{-1} \nonumber \\
          &= -\mK_c \mW^T \Psi^{-1}.
\end{align*}

We now have all the results to prove Theorem~\ref{theorem_appendix}.
\begin{proof}[\textbf{Proof of Theorem~\ref{theorem_appendix}}]
Comparing the errors corresponding to $\hat\vz_\beta$~\eqref{eq:e_beta} and $\hat\vz_c$~\eqref{eq:e_c}, we have 
\begin{align*}
    &e(\hat\vz_\beta) - e(\hat\vz_c) \\
    &= (\vz^T \mK_\beta^T \mK_\beta \vz + \Trace{\mG_\beta \Psi \mG_\beta^T}) \nonumber \\
    & \qquad \quad -  (\vz^T \mK_c^T \mK_c \vz + \Trace{\mG_c \Psi \mG_c^T}) \\
    &= \vz^T (\mK_\beta^T \mK_\beta - \mK_c ^T \mK_c) \vz \nonumber  + \Trace{\mG_\beta \Psi \mG_\beta^T} -  \Trace{\mG_c \Psi \mG_c^T} \\
    &= \vz^T (\mK_\beta^2 - \mK_c ^2) \vz \nonumber + \Trace{\Psi \mG_\beta^T\mG_\beta } -  \Trace{ \Psi \mG_c^T\mG_c} \\
    &= \underbrace{\vz^T (\mK_\beta - \mK_c)(\mK_\beta + \mK_c) \vz}_{\geq 0} \nonumber + \Trace{\Psi (\mG_\beta^T\mG_\beta - \mG_c^T\mG_c)} \\
    &\geq \Trace{\Psi (\mG_\beta^T\mG_\beta - \mG_c^T\mG_c)} \nonumber \\
    &= \Trace{\Psi\Big( \Psi^{-1} \mW (\mK_\beta^T \mK_\beta - \mK_c^T \mK_c) \mW^T \Psi^{-1} \Big)} \nonumber \\
    &=  \Trace{(\mK_\beta^T \mK_\beta- \mK_c^T \mK_c) \mW^T \Psi^{-1}\mW  } \nonumber \\
    &= \Trace{(\mK_\beta- \mK_c) (\mK_\beta + \mK_c) \mW^T \Psi^{-1}\mW  } \nonumber \\
    &\geq 0.
\end{align*}
In the derivation above, we used the fact that $(\mK_\beta - \mK_c )(\mK_\beta + \mK_c) \succeq 0$ since $\vz^T (\mK_\beta - \mK_c )(\mK_\beta + \mK_c) \vz \geq 0 \  \forall \ \vz$ and $(\mK_\beta - \mK_c )(\mK_\beta + \mK_c)$ is symmetric. Similarly, the matrix $(\mK_\beta- \mK_c) (\mK_\beta + \mK_c) \mW^T \Psi^{-1}\mW  \succeq 0$ and has non-negative eigenvalues, leading to a non-negative trace.
\end{proof}

\section{Normalized Hotelling deflation}\label{sec:normalized_hd}
The deflation scheme to generate higher-dimensional embeddings in our previous work~\citep{subramanian2021multimodal} is a normalized version of the Hotelling deflation scheme. The first step of this update scheme is
\begin{align}\label{normalized_hd_1}
    \mC_{j} &= \mC_{j-1} - \dfrac{\langle \mC_{j-1}, \vu_j \vv_j^T \rangle_v}{\|{\vu_j \vv_j^T}\|} \vu_j \vv_j^T, 
\end{align}
where $\langle . , . \rangle_v$ denotes the inner product of the vectorized version of the matrix entries.
Under unit-norm constraints on the canonical variates of the CCA problem, \eqref{normalized_hd_1} can be shown to reduce exactly to the Hotelling update, as following. Assume that the SVD decomposition for the matrix $\mC_1$ is given by $\mC_1 = \sum_{\ell=1}^r \sigma_\ell \boldsymbol{\alpha}_\ell \boldsymbol{\beta}_\ell^T$, where $r$ is the rank of the cross correlation matrix $\mC_1$. Further, it can be shown that the vectorized inner product $\langle \valpha_i \vbeta_i^T, \valpha_j \vbeta_j^T \rangle_v$ is equal to $\langle \valpha_i, \valpha_j \rangle \langle \vbeta_i, \vbeta_j \rangle$.

For $j=2$, we have
\begin{align*}
    \mC_2 &= \mC_1 - \dfrac{\langle \mC_1, \vu_1 \vv_1^T \rangle_v}{\|{\vu_1 \vv_1^T}\|} \vu_1 \vv_1^T \nonumber\\
    &=\mC_{1} - \dfrac{\langle \sum_{\ell=1}^r \rho_{\ell} \valpha_{\ell} \vbeta_{\ell}^T, \vu_1 \vv_1^T \rangle_v}{\|{\vu_1 \vv_1^T}\|} \vu_1 \vv_1^T, \nonumber \\
\end{align*}

which can be simplified to
\begin{align*}
    \mC_2 &= \mC_{1} - \dfrac{\langle \sum_{\ell=1}^r \rho_{\ell} \valpha_{\ell} \vbeta_{\ell}^T, \valpha_1 \vbeta_1^T \rangle_v}{\|\valpha_1 \vbeta_1^T\|} \valpha_1 \vbeta_1^T \nonumber  \\
    &= \mC_{1} - \dfrac{\langle \sum_{\ell=1}^r \rho_{\ell} \valpha_{\ell} \vbeta_{\ell}^T, \valpha_1 \vbeta_1^T \rangle_v}{\langle \valpha_1 \vbeta_1^T, \valpha_1 \vbeta_1^T \rangle_v^{1/2}} \valpha_1 \vbeta_1^T  \nonumber \\
    &= \mC_{1} - \rho_{1} \dfrac{ \langle \valpha_{1} \vbeta_{1}^T, \valpha_1 \vbeta_1^T \rangle_v}{\langle \valpha_1 \vbeta_1^T, \valpha_1 \vbeta_1^T \rangle_v^{1/2}} \valpha_1 \vbeta_1^T  \nonumber \\
    &= \mC_{1} - \rho_{1} \langle \valpha_{1} \vbeta_{1}^T, \valpha_1 \vbeta_1^T \rangle_v^{1/2} \valpha_1 \vbeta_1^T  \nonumber  \\
    &= \mC_{1} - \rho_{1} \langle \valpha_{1}, \valpha_1 \rangle^{1/2} \langle \vbeta_1, \vbeta_1 \rangle^{1/2} \valpha_1 \vbeta_1^T  \nonumber \\
    &= \mC_{1} - \rho_1 \valpha_1 \vbeta_1^T \nonumber \\
    &=  \sum_{\ell=2}^r \rho_{\ell} \valpha_{\ell} \vbeta_{\ell}^T,
\end{align*}
which is equivalent to the Hotelling update for this setting. This can be extended to further iterations. Let's assume that at any iteration $i$, we have $\mC_i = \sum_{\ell=i}^r \rho_{\ell} \valpha_{\ell} \vbeta_{\ell}^T$. Then, at iteration $i+1$,
\begin{align*}
    \mC_{i+1} &= \mC_{i} - \dfrac{\langle \mC_i, \vu_i \vv_i^T \rangle_v}{\|{\vu_i \vv_i^T}\|} \vu_i \vv_i^T \nonumber \\
    &= \mC_{i} - \dfrac{\langle \sum_{\ell=i}^r \rho_{\ell} \valpha_{\ell} \vbeta_{\ell}^T, \valpha_i \vbeta_i^T \rangle_v}{\|\valpha_i \vbeta_i^T\|} \valpha_i \vbeta_i^T \nonumber  \\
    &= \mC_{i} - \rho_{i} \langle \valpha_{i} \vbeta_{i}^T, \valpha_i \vbeta_i^T \rangle_v^{1/2} \valpha_i \vbeta_i^T  \nonumber  \\
    &= \mC_{i} - \rho_{i} \langle \valpha_{i}, \valpha_i \rangle^{1/2} \langle \vbeta_i, \vbeta_i \rangle^{1/2} \valpha_i \vbeta_i^T \nonumber \\
    &= \sum_{\ell={i+1}}^r \rho_{\ell} \valpha_{\ell} \vbeta_{\ell}^T,
\end{align*}
such that the first step of the update scheme in~\cite{subramanian2021multimodal} is the same as the Hotelling deflation in this setting. The second step of the update scheme involving the scaling normalization of the cross-covariance matrix is primarily for numerical reasons. Therefore, the overall update scheme in~\cite{subramanian2021multimodal} is a normalized version of the Hotelling deflation. The scale factor will propagate across the iterations, such that the properties of the Hotelling deflation would continue to hold here.

\bibliographystyle{IEEEtran}
\bibliography{references}

\begin{IEEEbiographynophoto}{Vaishnavi Subramanian}
Vaishnavi Subramanian is a PhD candidate in Electrical and Computer Engineering 
department at Coordinated Science Lab at the University of Illinois at 
Urbana-Champaign (UIUC). She completed her M.S. degree from UIUC in 2018 and her B.Tech in Electrical Engineering from IIT Bombay in 2016. Her current research interests include signal processing, machine learning, bioinformatics, health informatics, image processing, medical data processing, multi-modality data analysis, and interpretable prediction systems. Vaishnavi has served in the 
committee for the CSL Student Conference 2019  and 2020  held at UIUC.  
She is the recipient of the Mavis Future Faculty Fellowship, and the Elsa 
and Floyd Dunn Award for biomedical research. 
\end{IEEEbiographynophoto}

\begin{IEEEbiographynophoto}{Tanveer Syeda-Mahmood}
Dr. Tanveer Syeda-Mahmood is an IBM Fellow and Global Imaging AI Leader in IBM Research.  Previously, she was the Chief Scientist/overall lead for the Medical Sieve Radiology Grand Challenge project in IBM Research, Almaden. Dr. Syeda-Mahmood graduated from the MIT AI Lab in 1993 with a Ph.D in Computer Science. Over the past 30 years, her research interests have been in a variety of areas relating to artificial intelligence including computer vision, image and video databases, medical image analysis, bioinformatics, signal processing, document analysis, and distributed computing frameworks. She has over 250 refereed publications and over 120 patents filed. Dr. Syeda-Mahmood is the General co-Chair of MICCAI 2023, the premier conference in medical imaging. She is also the Program co-Chair of IEEE ISBI 2022 to be held in Calcutta, India. Dr. Syeda-Mahmood is a Fellow of IEEE.  
\end{IEEEbiographynophoto}

\begin{IEEEbiographynophoto}{Minh N. Do}
Minh N. Do was born in Thanh Hoa, Vietnam in 1974. He received the B.Eng. degree in Computer Engineering from the University of Canberra, Australia in 1997, and the Dr.Sci. degree in Communication Systems from the Swiss Federal Institute of Technology Lausanne (EPFL), Switzerland in 2001.
Since 2002, he has been on the faculty at the University of Illinois at Urbana-Champaign, where he is currently the Thomas and Margaret Huang Endowed Professor in Signal Processing \& Data Science in the Department of Electrical and Computer Engineering, and holds affiliate appointments with the Coordinated Science Laboratory, the Beckman Institute for Advanced Science and Technology, the Department of Bioengineering, and the Department of Computer Science. 
His current research interests include signal processing, computational imaging, machine perception, and data science. 
\end{IEEEbiographynophoto}




\end{document}